\documentclass[english]{article}
\usepackage{PRIMEarxiv}

\usepackage[utf8]{inputenc} 
\usepackage[T1]{fontenc}    
\usepackage{hyperref}       

\usepackage{color}
\usepackage{babel}
\usepackage{array}
\usepackage{rotfloat}
\usepackage{multirow}
\usepackage{algorithmic}
\usepackage{algorithm}
\usepackage{amsmath}
\usepackage{amsthm}
\usepackage{amssymb}
\usepackage{graphicx}
\usepackage{setspace}
\usepackage{acronym}
\usepackage{geometry}
\usepackage{color}

\DeclareMathOperator*{\argmax}{arg\,max}

\title{Machine learning-based patient selection in an emergency department}                      

\author{
  Furian, Nikolaus \\
  Institute of Engineering and Business Informatics \\
  Graz University of Technology \\
  Graz, Austria\\
  \texttt{nikolaus.furian@tugraz.at} \\
   \And
  O'Sullivan, Michael \\
  Department of Engineering Science \\
  University of Auckland \\
  Auckland, New Zealand\\
  \texttt{michael.osullivan@auckland.ac.nz} \\
  
  \And
  Walker, Cameron \\
  Department of Engineering Science \\
  University of Auckland \\
  Auckland, New Zealand\\
  \texttt{cameron.walker@auckland.ac.nz} \\
  
  \And
  Reuter-Oppermann, Melanie \\
  Information Systems \\
  Technical University of Darmstadt \\
  Darmstadt, Germany\\
  \texttt{melanie.reuter-oppermann@tu-darmstadt.de} \\
  
}

\acrodef{ED}{Emergency Department}
\acrodef{ML}{Machine Learning}
\acrodef{APQ}{Accumulated Priority Queuing}
\acrodef{LOS}{Length of Stay}
\acrodef{KPI}{Key Performance Indicator}
\acrodef{TTD}{Time to Doctor}
\acrodef{ESI}{Emergency Severity Index}
\begin{document}
\maketitle

\begin{abstract}
The performance of \acp{ED} is of great importance for any health care system, as they serve as the entry point for many patients. However, among other factors, the variability of patient acuity levels  and corresponding treatment requirements of patients visiting \acp{ED} imposes significant challenges on decision makers. Balancing waiting times of patients to be first seen by a physician with the overall length of stay over all acuity levels is crucial to maintain an acceptable level of operational performance for all patients. To address those requirements when  assigning idle resources to patients, several methods have been proposed in the past, including the \ac{APQ} method. The \ac{APQ} method linearly assigns priority scores to patients with respect to their time in the system and acuity level. Hence, selection decisions are based on a simple system representation that is used as an input for a selection function. This paper investigates the potential of an \ac{ML} based patient selection method. It assumes that for a large set of training data, including a multitude of different system states, (near) optimal assignments can be computed by a (heuristic) optimizer, with respect to a chosen performance metric, and aims to imitate such optimal behavior when applied to new situations. Thereby, it incorporates a comprehensive state representation of the system and a complex non-linear selection function. The motivation for the proposed approach is that high quality selection decisions may depend on a variety of factors describing the current state of the \ac{ED}, not limited to waiting times, which can be captured and utilized by the \ac{ML} model. Results show that the proposed method significantly outperforms the \ac{APQ} method for a majority of evaluated settings.
\end{abstract}

\keywords{OR in health services \and Emergency Department \and Machine Learning \and Accumulated Priority Queuing}

\section{Introduction}

\acfp{ED} play a crucial role in health care systems worldwide, as they serve as the entry point for many patients, witnessing highly stochastic patient arrival patterns while operating 24 hours each day of the year. However, many health care providers face major issues regarding the operational performance of \acp{ED}, such as overcrowding, diversion, or long and imbalanced waiting times for patients with varying priorities. The reasons for those inefficiencies are manifold and include among others: budget restrictions; the aging of populations; insufficient resources; or non urgent patient visits ( \cite{Hoot2008,Moskop2009}).

As a consequence, improving \ac{ED} operations by the design of new care delivery processes or adequate allocation of resources has gained significant interest for both health care stakeholders and researchers, over the last decades \cite{Furian2018}. Proposed approaches include: virtual streams \cite{Saghafian2012}; fast track units \cite{Paul2012}; or advanced patient and resource selection methods such as \acf{APQ} \cite{Stanford2014, Ferrand2018, Cildoz2019}. Modeling and simulation has been extensively used by decision makers to evaluate the performance of these methods within a virtual abstraction of the system, before actually implementing them in a real world \ac{ED}.

In particular, some recent studies, e.g. \cite{Cildoz2019} or \cite{Ferrand2018}, argue that \ac{APQ} is a technique capable of improving the performance of \acp{ED} significantly. By the classification of Furian et al. \cite{Furian2018}, \ac{APQ} is a patient selection policy, i.e. a method to select the next patient from a pool of waiting patients to be assigned to an idle resource, e.g. a doctor, for treatment or consultation. The main aim of the \ac{APQ} method is to balance performance measures for patients belonging to different priority categories and thereby to improve an overall objective function, or measure, representing the operational performance of the entire \ac{ED}. While traditional queuing policies for \acp{ED} select patients purely by their priorities (i.e. high priority patients are always seen before low priority patients), and type of consultation (i.e. first or subsequent visit of doctors), \ac{APQ} further considers the time patients are already present in the \ac{ED}. Therefore, patients are separated in virtual queues of priority grades and type of consultation and each queue is assigned a (fixed) weight parameter. When selecting a patient, a score value is computed for each patient being at the head of one of the queues. The score is obtained by the product of the weight parameter of the corresponding queue and the time the patient already spent in the system. The patient with the highest score is then seen next by an idle resource.

In its essence, \ac{APQ} can be abstracted in the following way. Whenever a resource can be dispatched, a simple representation of the current state of the \ac{ED} is computed and passed to a mathematical function for selecting a waiting patient. The time spent in the system by a patient that is currently at the head of a virtual queue defines the system's state, and the function applies simple linear weight functions to each of those patients. Therefore, the weight parameters for queues are chosen, e.g. via simulation optimization, such that a pre-defined objective function, accounting for the performance of the studied \ac{ED}, is optimized over a training data set.

While the simplicity of the approach seems to be beneficial for practical reasons, the question whether more comprehensive state representations and complex selection functions may further improve the performance of an \ac{ED} has not been considered yet. In other words, can considering additional information on the state of the \ac{ED} (such as queue lengths, time of the day, number of patients that are close to target times, etc.) and possibly non-linear functions mapping those states to selection decisions result in improved performance measures?

In theory, for every possible state of a specific \ac{ED} there is at least one optimal decision to select a patient to be seen. Obviously, these optimal decision are not only dependent on the state of the \ac{ED} at the time of decision making, but also on future events (e.g. arrival of new patients, finishing of ongoing consultations and treatments, future requirements of patients, etc.) that may occur in the system. As systems like \acp{ED} are of a highly stochastic nature, it is impossible to know, or even accurately predict, those future events at the time of decision making. However, having a virtual abstraction of the system, i.e. a model including all stochastic aspects of the \ac{ED}, allows for the generation (or sampling) of (in theory) unlimited realizations of the \ac{ED} for a chosen period of time. For example, it is possible to sample the arrival process of patients, e.g. the realization of a stochastic process, and treatment requirements and times, e.g. the realization of random variables, of all patients arriving at the \ac{ED} during a single day. For such a realization, from now on referred to as an instance, patient related information could be treated as deterministic data. 

Further, assuming to have perfect and full information on such an instance allows to treat the patient selection problem as a scheduling problem. In other words, it is possible to compute an optimal plan with respect to a chosen objective function, or schedule, determining at what time which patient is seen by which resource, considering all patient requirements and events throughout the entire time span covered by the instance. Obviously, such a single optimal schedule is of no practical use, as perfect and complete information on future events will never be available in a real world setting. However, computing optimal solutions for a large set of instances for a specific \ac{ED} enables the definition, fitting and use of complex functions to select patients at a given point in time - an approach that is analogous to the use of simulation optimization to find optimal weights for the \ac{APQ} policy.

The \acf{ML} based patient selection method presented in this paper uses such optimal plans for a set of training instances to fit a (non-linear) \ac{ML} model to learn optimal selection decisions with respect to a comprehensive state representation of the \ac{ED}. In other words, given a state of the \ac{ED} it predicts the selection of a patient that potentially leads to an optimal, or at least high quality, decision. Thereby, the transformation of optimal solutions for training instances to an \ac{ML} model may be seen as building a computational ``experience'' (or knowledge base) that enables the selection of patients with respect to the state of the \ac{ED} at the time of decision making, and implicitly possible future events in the system. 

A discrete event simulation model of a typical \ac{ED} is proposed in order to generate training instances, evaluate the approach on new unseen instances and compare it with results obtained by standard methods from literature, e.g. \ac{APQ} patient selection. The model covers most relevant aspects of \ac{ED} operations and includes stochastic elements (e.g. stochastic processes for patient streams, stochastic treatment times, etc.) to account for the dynamic nature of \acp{ED}. Results show that the \ac{ML} based patient selection method significantly outperforms standard approaches for a majority of proposed scenarios and definitions of objective functions. 

The paper is structured as follows: in the next section relevant literature and the contribution of the paper are outlined. In Section \ref{sec::EDModel}, the simulation model used in this study is introduced, a generic and parameterizable objective function is defined, and standard patient selection methods are described. Section \ref{sec::MLPatientSelection} introduces the \ac{ML}-based patient selection method. Input parameters and scenarios are defined in Section \ref{sec::DataAndScenarios}. Results are presented in Section \ref{sec::Results}. The paper concludes with final remarks and suggestions for further research in Section \ref{sec::Conclusion}. 

\section{Related Work and Contribution} \label{sec::RelatedWork}

The analysis and improvement of \ac{ED} operations has been addressed by many researchers in the past decades. Thereby, modeling and simulation has been extensively used as a decision support tool by researchers and stakeholders. As a consequence, there exists a vast number of publications on modeling of \ac{ED} systems. For comprehensive reviews on \ac{ED} modeling and simulation, the reader is referred to \cite{Saghafian2015} and \cite{Gul2015}. A survey on simulation models dealing with overcrowding effects in \acp{ED} is given by \cite{Paul2010}. 

In a previous paper, the authors propose a guideline for \ac{ED} modeling, that is based on generic building blocks of \ac{ED} operations, see \cite{Furian2018}. In addition, a variety of patient selection methods integrated in, and evaluated by, \ac{ED} simulation models are identified.

A majority of proposed models select patients by a purely priority-based policy. In particular, patients are selected for consultation and/or treatment based on their previously assigned triage grades. Patients with equivalent triage grades are then selected based on the `first-in-first-out'' (FIFO) principle. Examples for such selection policies are given, among others, by \cite{Ahmed2009,Ashour2013,Bair2010,Oh2016,Facchin2010,Lin2015}.

A common extension to purely priority based selection policies is to further classify patients with respect to the number of times they have been seen by a doctor, i.e. new vs. old patients, and define rules considering those classifications. Such approaches were proposed by \cite{Chonde2013,Facchin2010, Ghanes2014,Saghafian2012,Tan2012,Vrugt2016,Yang2016}.

In addition to the number of visits, \cite{Saghafian2012} classify patients into virtual streams with respect to an estimation of a patient being discharged or admitted to the hospital at the end of the stay in the \ac{ED}. The reasoning behind this policy is that conflicting performance measures for patients in either class may be of interest. Whereas for patients being discharged at the end of their stay in the \ac{ED}, the \ac{LOS} is mainly of interest, the \ac{TTD} may be more important for patients with a high probability of being admitted to the hospital.

The work of \cite{Eskandari2011} includes also non-\ac{ED} patients being seen in diagnostic facilities, and proposes a triage based selection policy for both patients categories.

Less work has been published on policies that include resource related information in patient selection policies, referred to as ``combined patient and resource selection'' by \cite{Furian2018}. For example, \cite{Conelly2004} introduce a policy that aims to balance the share of currently seen high and low priority patients for each doctor individually. A policy including skill and/or training levels of doctors when selecting a patient is proposed by \cite{Hay2006}.

A two step patient selection policy based on a queuing network formulation (rather than a simulation model) of an \ac{ED} is introduced by \cite{Huang2015}. They classify patients between triage patient (TP) (i.e. first contact with a medical staff member) and patients in process (IP). Whenever a physician becomes idle, it is first determined whether an IP or TP patient is seen. This first step always selects TPs, if any exist and are close to reach a predefined time limit for getting triaged. In the second step, given that such TPs exist, the patient whose time in the system is closest to the limit is selected. If a TP is to be selected in the second step, the policy uses a modified $c\mu$-rule to decide which patient to see next. 

A queuing principle for \ac{ED} systems that recently gained significant attention is \ac{APQ}. For a general discussion of the \ac{APQ} method, i.e. without a focus on \ac{ED} operations, the reader is referred to \cite{Kleinrock1964} or \cite{Stanford2014}. However, those general models do not include common features of \acp{ED}, such as patients being seen multiple times by doctors or time-dependent arrival rates of patients. \cite{Hay2006} were among the first to propose a policy for \ac{ED} patient selection that can be considered to follow an \ac{APQ} principle. They assign patients an ``operating priority'' that is composed by a waiting time component (that increases over time) and a ``clinical priority'' (similar to triage grades). \cite{Mes2012} state that patients are prioritized with respect to a function of triage categories and waiting times that increases the priority the waiting time of a patient is close to target. However, they do not exactly specify the function that is used. Dynamically increasing priorities of patients are also proposed by \cite{Tan2012}. In particular, they define three priority functions. The first assigns priorities based on an exponential function of the estimated consultation times of patients. The second uses an exponential function on the remaining time of a patient until a \ac{LOS} target is met, and a linear weight function when for patients that already missed the target. The third policy combines those by a linear weighted sum.

In a more recent study, \cite{Ferrand2018} discuss the performance of the \ac{APQ} method by a simulation model based on a real world \ac{ED}. The authors show that \ac{APQ} outperforms other strategies, such as fast track units or virtual streaming, to balance \ac{KPI} over patients with different triage grades. However, their results are based on the \ac{LOS} as a performance measure, and hence do not include an investigation on how \ac{APQ} performs with respect to target related \acp{KPI}.

The work of \cite{Cildoz2019} addresses this shortcoming, and defines a performance measure based on a weighted sum of the total waiting time of patients, and the share of patients meeting a \ac{TTD} target time. Besides a strictly linear \ac{APQ} model, the authors also propose a so called \ac{APQ}-h model that prevents priorities to increase over time as soon as the \ac{TTD} target time is met by a patient. However, they conclude that the difference in performance of both policies is not significant for tested scenarios. They evaluate the \ac{APQ} under a variety of settings, including different patient mixes, arrival rates and resource utilization rates.

The use or application of \ac{ML} methods and tools to analyse and/or improve health care systems has gained increased attention by practitioners and the scientific community. Reuter-Oppermann and K\"{u}hl review publications that use machine learning approaches either to directly address planning problems in healthcare logistics or to derive input for these \cite{Reuter2021}. They state that a significant share of publications has targeted emergency departments.
To the best of our knowledge, only few papers have used machine learning within ED optimisation or simulation. 
Laskowski, for example, combines an agent based model with a machine learning approach to build a hybrid system for decision support in order to improve processes in an emergency department \cite{laskowski}.
Ceglowski et al. integrate a data-mining approach into a discrete-event simulation in order to analyse ED processes and identify bottlenecks in the interface between the ED and the hospital wards \cite{ceglowski}.

Significantly more publications have used machine learning approaches to predict arrival rates \cite{acid,afilal,menke,xu}, patient admissions \cite{graham,kramer,lee2019}, lengths of stay \cite{benbelkacem,cai} or patient re-admissions \cite{turgeman,zheng, lee,futoma,rana,eigner,nilmini}.

Menke et al. and Xu et al. for example use artificial neural networks (ANN) to predict the daily patient volumes arriving at an ED \cite{menke,xu}, while Xu et al. compare the ANN with the nonlinear least square regression (NLLSR) and the multiple linear regression (MLR) \cite{xu}. 
Afilal et al. apply time-series based forecasting models to predict longterm as well as short term arrival rates \cite{afilal}.

Graham et al. compare logistic regression, decision trees and gradient boosted machines (GBM) to predict which patients are admitted into the hospital from the ED \cite{graham}. Kr\"{a}mer et al. use supervised machine learning techniques to classify hospital admissions into urgent and elective care \cite{kramer}, while Lee et al. \cite{lee2019} distinguish four admission classes, i.e. intensive care unit, telemetry unit, general practice unit, and observation unit, for the ML-based classification approaches.

Patient readmissions are addressed by Turgeman and May with a support vector machine (SVM) \cite{turgeman} and by Zheng et al. with classification models that use neural networks, random forest and support vector machines, with a specific focus on heart failure patients \cite{zheng}. Lee et al. have developed a clinical tool based on machine learning for predicting patients who will return within 72 hours to the pediatric emergency department \cite{lee}. Futoma et al. compare predictive and deep learning models for 30 day re-admissions \cite{futoma}.

The integration of classical OR methods, such as optimization and simulation, with \ac{ML} models, has gained significant interest within the scientific community in recent years. For a comprehensive guide on the combined use of those methodologies, the reader is referred to \cite{Bengio2021}. Due to the novelty of the field, most published studies focus on classical combinatorial optimization problems, such as the Traveling Salesmen problem, e.g. \cite{Bello2016} and \cite{Joshi2019}, variants of the vehicle routing problem, e.g. \cite{Nazari2018} and \cite{Furian2021}, or the facility location problem, e.g. \cite{Lodi2020}. However, some studies propose the integrated use of classical optimization with \ac{ML} models in more practical settings, e.g. \cite{Hottung2020} introduce a learning assisted tree search for a container pre-marshalling problem and \cite{Fischetti2019} predict optimal objective values of an offshore wind-park production optimization problem.  

In general, \cite{Bengio2021} consider three ways to integrate \ac{ML} into optimization tasks: (1) end-to-end learning, i.e. heuristically computing solutions for the optimization problem via an \ac{ML} model; (2) using \ac{ML} models to  to gather information on the problem at hand in a pre-processing step and pass that information on to a classical optimization algorithm; and (3) utilize \ac{ML} models to to make online heuristic decisions within classical optimization algorithms.

The proposed \ac{ML}-based patient selection policy does not strictly fall in any of those categories. Although the method uses an \ac{ML} model to imitate optimal behavior, i.e. optimized schedules of patients, and hence may be categorized as end-to-end learning, it is used as a policy rather than an algorithm. In particular, one does not solve an optimization problem in the classical sense when selecting a patient for consultation.

To the best knowledge of the authors, the presented approach is the first using \ac{ML} models to select patients for consultation/treatment in a healthcare setting, and in particular for \acp{ED}. While previously published methods rely on relatively simple state representations of an \acp{ED}, our method includes a comprehensive abstraction of an \ac{ED}'s state for decision making. Further, the use of deterministic optimization results as a knowledge base for deriving selection policies has not been considered in previous research.     

\section{The \ac{ED} Setting} \label{sec::EDModel}

The \ac{ED} model proposed in this paper is not the abstraction of a specific real world \ac{ED}, but is rather intended to be a representation of ``typical'' \ac{ED} operations. In a previous study \cite{Furian2018}, the authors analyzed and classified a wide range of published \ac{ED} models and defined archetype building blocks of the latter. The presented model is a composition of the most commonly used building blocks, balancing implied simplifications and assumptions with the requirement to represent generic and ``typical'' \ac{ED} processes and resource definitions. For example, the register and triage process of the proposed model has been used by 18 out of 48 models analyzed in \cite{Furian2018}. The patient treatment processes in the model are a realizations of the generic $P_{P,0}[A_{A,0},D_{D,0}]T_{T,0}$ process classified by \cite{Furian2018} that covers about 60\% of analyzed models. For detailed information on the nomenclature the interested reader is referred to \cite{Furian2018}. In the following, the components of the model are outlined in detail.

\begin{table}[ht!]
	\begin{center}
		\begin{tabular}{l l }
			\hline
			 Definition & Description \\
			 \hline
			 \multicolumn{2}{l}{General Definitions} \\
			 \hline
			 $J= \{ 2,3,4,5 \}$ & Set representing triage grades \\
			 $P = \bigcup\limits_{j\in J} P_j$ & Set of patients, partitioned over triage grades \\
			 $p_j$ & Probability of a patient being of class $P_j$ \\
			 $pd_j$ & Probability of a patient of class $j$ is requiring diagnostic tests \\
			 $pa_j$ & Probability of a patient of class $j$ is arriving via ambulance \\
			 $pw_j$ & Importance weight associated with patient class $j$ \\
			 $D$ & Set of doctors \\
			 $\lambda_t$ & Arrival rate of patients at time $t$ of a given day \\
			\hline
			\multicolumn{2}{l}{General Time Related Definitions} \\
			 \hline
			 $t_a^i, i \in P$ & Arrival time of patient $i$ \\
			 $t_l^i, i \in P$ & Discharging time of patient $i$ \\
			 $t_{fc}^i, i \in P$ & Start time of first consultation of patient $i$ \\
			 $t_{sc}^i, i \in P$ & Start time of second consultation of patient $i$ \\
			 $ttd^i = t_{fc}^i - t_a^i, i \in P$ & Time to doctor of patient $i$ \\
			 $los^i = t_{l}^i - t_a^i, i \in P$ & Length of stay of patient $i$ \\
			 $t_{ttd}^j, j \in J$ & Maximum time targets (in minutes) for $ttd$ over triage grades \\
			 $ts_{ttd}^j, j \in J$ & Target share of patients that should meet \ac{TTD} target times \\
			 $t_{sw_3}^i, i \in P$ & Start of waiting time $w_3$ of patient $i$  \\
			 $t_{sw_4}^i, i \in P$ & Start of waiting time $w_4$ of patient $i$  \\
			 \hline
			    \multicolumn{2}{l}{Waiting Time Related Definitions} \\
			 \hline
			 $w_1^i, i \in P $ & Waiting time for register of patient $i$ \\
			 $w_2^i, i \in P $ & Waiting time for triage of patient $i$ \\
			 $w_3^i, i \in P $ & Waiting time for first consultation of patient $i$ \\
			 $w_4^i, i \in P $ & Waiting time for second consultation of patient $i$ \\
			 $W_1^i, i \in P$ & Total waiting time until first consultation of patient $i$ \\
			 $W^i, i \in P$ & Total waiting time of patient $i$ \\
			 \hline
			    \multicolumn{2}{l}{Queuing Related Definitions} \\
			 \hline
			 $Q_{j,1}, j \in J$ & patients of category $j$ waiting for the first consultation \\
			 $Q_{j,2}^d, j \in J, d \in D$ & patients of category $j$ waiting for the second consultation with doctor $d$ \\
			 \hline
		\end{tabular}
		\caption{Summary of Definitions}
		\label{tab::SummaryParameter}
	\end{center}
\end{table}

\subsection{Patient Categories and Processes} \label{sec::PatientProcess}

A standard method to categorize patients are triage grades which reflect the severity of a patient's condition upon arrival at the \ac{ED}. Most commonly used triage grading systems include the \ac{ESI} and the Manchester Triage Scale (MTS). Both consist of 5 triage levels, ranging from "immediately life threatening" to "not urgent" conditions of patients. Note, that the \ac{ESI} not only accounts for a patient's condition, but also for the expected amount of resources required. The presented model is also based on a 5 level triage grading system, where triage grade 1 is assigned to resuscitation patients and triage grade 5 to the least urgent patients. 

Since, resuscitation patients are usually admitted to trauma bays or resuscitation facilities, and hence do not compete for resources with other patients, the included triage grades range from 2 (high priority) to 5 (low priority), similar to \cite{Cildoz2019}.

Besides triage grades, the mode of arrival is another criteria to classify patients. Patients either arrive by ambulance or are classified as ``walk-in'' patients, i.e. arriving by their own means. However, we assume that the model of arrival affects the care process only for patients with triage grade 2. Those are assumed to be triaged on the ambulance and therefore skip the first process steps, as will be described below.

Upon arrival, patients wait to be registered at the \ac{ED}. The registration itself is followed by a wait for a triage activity. As mentioned above, category 2 patients arriving by ambulance skip those two activities. Further, category 2 patients arriving by a private mode of transportation, i.e. walk-in patients, are assumed to skip the registration step and start their process with a short triage activity. After triage (or arrival of category 2 patients by ambulance) patients queue up to be first seen by a doctor, referred to as ``first consultation''. 

This first consultation is possibly followed by some diagnostic tests. The probability of requiring such tests is assumed to be triage category dependent. However, no explicit resources for performing those tests are included in the model, i.e. they are modeled as a (stochastic) time delay. We assume that (if required) diagnostic tests start immediately after the first consultation, and consequently any possible waiting time is included in the time delay for testing. Patients not requiring diagnostic tests are assumed to leave the \ac{ED} after the first consultation. 

After the diagnostic tests, patients queue up for the ``second consultation'' by a doctor, which may be seen as the final treatment. The resulting waiting time (denoted by $w_4$) starts immediately after the time delay for testing. It is important to note that patients are required to be seen by the same doctor for the second consultation who performed the first consultation. This policy is often referred to as ``same-doctor-policy'' and is commonly found in literature, see for example \cite{Saghafian2012,Ghanes2014,Konrad2013}.

\begin{figure}[ht!]
    \begin{center}
    \includegraphics[width=16cm]{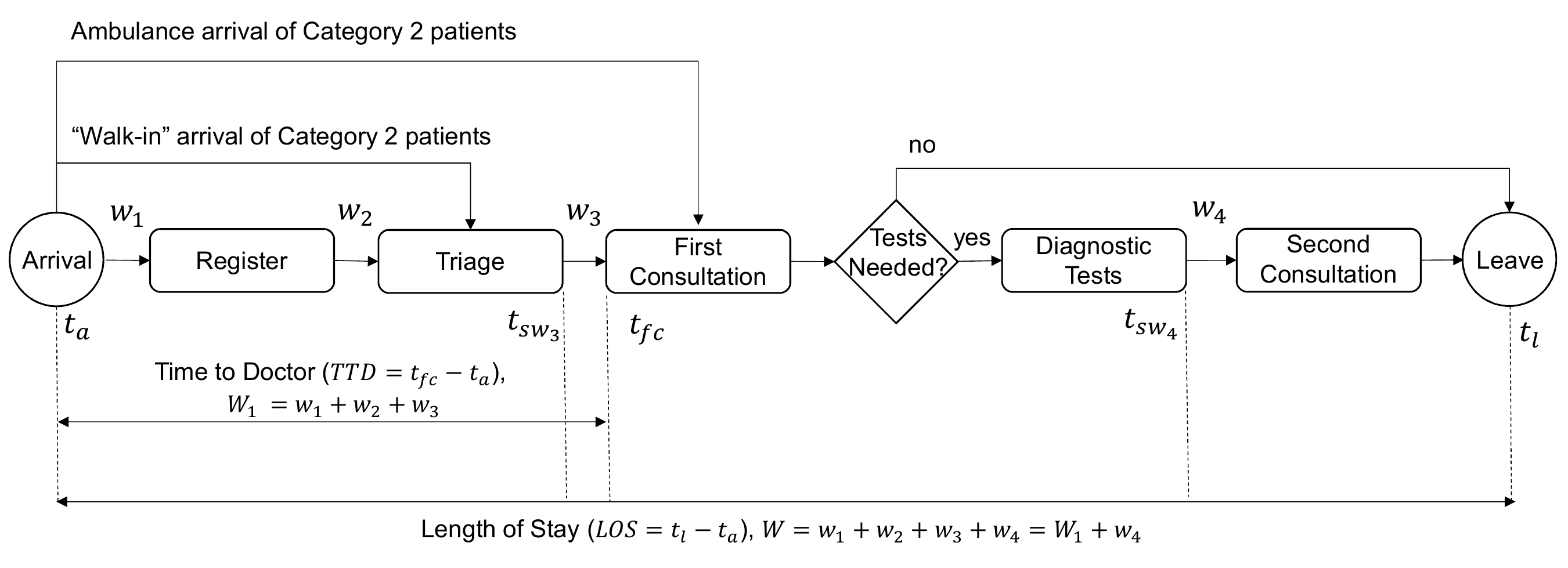}
    \caption{\ac{ED} Patient Process}
    \label{fig:PatientProcess}
    \end{center}
    
\end{figure}

Table \ref{tab::SummaryParameter} summarizes relevant formal definitions to describe the system, while figure \ref{fig:PatientProcess} illustrates the patient process and important measures.

\subsection{Resources}

Patients require a set of resources along their care process. In the presented model, only resources which patients compete for, and which are not directly coupled to other resources, are included. For example, patients register with a register clerk, who is modeled as an explicit resource. A register facility, such as a registration desk, is not explicitly included, as we assume that whenever the clerk is available for registration, the associated physical facility is also available.

Further, we include a set of triage nurses, whose sole responsibility in the model is to perform triage activities of arriving patients. In particular, they do not participate in any consultation or test activities. For the first and second consultation we include a set of doctors (or physicians) $D$ that perform those activities. As stated in section \ref{sec::PatientProcess}, we do not include resources for diagnostic tests, but rather model them as stochastic time delays. Such a simplification is commonly used in \ac{ED} models, see for example \cite{Gunal2006,Paul2012, Lim2013, Vrugt2016}.

\subsection{Performance Metrics}

A wide range of performance indicators to evaluate \ac{ED} operations are used and reported in literature. Those range from simple time-based, such as \ac{LOS} or waiting times, to complex combined measures, such as the National Emergency Department Crowding Scale (NEDCOS). In a previous paper \cite{Furian2018}, the authors classified commonly used \acp{KPI} as: time-based, limit-based, state-based, financial measures, and combined measures. Time-based measures denote for \acp{KPI} that capture time spans along patient processes, such as \ac{LOS}, \ac{TTD}, or waiting times. Limit-based \acp{KPI} usually set time-based measures in context with predefined targets, for example measure the share of patients meeting a \ac{TTD} target time. State-based \acp{KPI} aggregate system states to a performance indicator, e.g. resource utilization rates or diversion state rates. Combined measures, such as the NEDCOS, \cite{Weiss2004}, aggregate a variety of single measures to an overall \ac{KPI}.

In this paper, we use different performance measures that can all be represented by a weighted sum of two individual indicators, i.e. are different parametrization of a combined \ac{KPI}. In the following, we will describe the individual components and define the analytical as well as the empirical performance metric. 

Therefore, let $X_{\tilde{W}}^j$ be a random variable representing the total waiting time for patients of category $j$ that need two consultations  and respectively $X_{\tilde{W}_1}^j$ be a random variable representing the total waiting time for patients of category $j$ that leave the \ac{ED} after the first consultation. Further, let $x^j_{i,ttd}$ be a variable that is $1$ if patient $i$ of category $j$ meets the \ac{TTD} target time, i.e. if $ttd^i \leq ttd^j$, and $0$ otherwise, and $X_{ttd}^j$ its random variable representation.

The analytical combined measure is then defined by

\begin{equation}
   f_{ED}(P|\lambda_{W}, \lambda_{ttd}) =\sum_{j \in J} p_j pw_j \left(\lambda_{W} \left((1-pd_j)E(X_{\tilde{W}_1}^j) + pd_j E(X_{\tilde{W}}^j) \right) + \lambda_{ttd} \max\left( ts^j_{ttd} - E(X_{ttd}^j),0\right) \right),
\end{equation}

where $\lambda_{W}$ and $\lambda_{ttd}$ are the corresponding weights of individual measures. In particular, $f_{ED}$ is a weighted sum of the expected total waiting time of patients (per category and diagnostic requirement) and the share of patients that are not meeting the \ac{TTD} target (per patient category).

Obviously, the real distributions of $X_{\tilde{W}}^j$, $X_{\tilde{W}_1}^j$, and $X_{ttd}^j$ are not known, but for the evaluation of a simulation result (or a data record) of $P$ patients, the following estimator can be used:

\begin{equation} \label{eq::EmpObj}
    \hat{f}_{ED}(P|\lambda_{W}, \lambda_{ttd}) =\sum_{j \in J} p_j pw_j  \left( \lambda_{W} \frac{1}{|P_j|} \sum_{i \in P_j} W^i + \lambda_{ttd} \left( ts^j_{ttd} - \frac{1}{|P_j|} \sum_{i \in P_j} x^j_{i,ttd}\right)^+ \right).
\end{equation}

The weight tuple $(\lambda_W, \lambda_{ttd})$ is used to generate a set of measures to evaluate the proposed method. Thereby, the goal is not to define a universally applicable measure, but to demonstrate the performance of the approach with respect to a variety of settings and goal functions.

\subsection{Patient Selection Policies and Constraints} \label{sec::PatientSelectionPolicies}

Besides entities and process definitions, the control logic or control policies determine the content of a model \cite{Furian2015}. For the model proposed in this paper, the control policies are concerned with matching available resources with patients queuing up for services. Those can be further categorized in resource, patient and combined selection policies, see \cite{Furian2018}. As this study is mainly concerned with the selection of patients, those polices are described in detail in the remainder of the section.

In general, patients queue for registration, triage, first and second consultation. However, as a patient's triage grade is not known before queuing up for the first consultation, i.e. before the completion of the triage activity or the arrival of a patient in case of ambulance arrivals of category $2$ patients, the queuing policies for registration and triage are modeled by the ``first-in-first-out'' (FIFO) principle.

In its most general form, the selection of patients to be seen by a doctor (for first and second consultation) at a given point in time can be seen as function that maps the set of available doctors to a sub-set of waiting patients. Such a function does not necessarily require that the set of waiting patients is structured by a system of queues. However, due to practical implications, such as constraints on the patients to be selected and the priority of patients imposed by the triage grades, most commonly used patient selection policies are based on a set of virtual queues that structure the set of waiting patients.

\begin{figure}[ht!]
    \begin{center}
    \includegraphics[width=10cm]{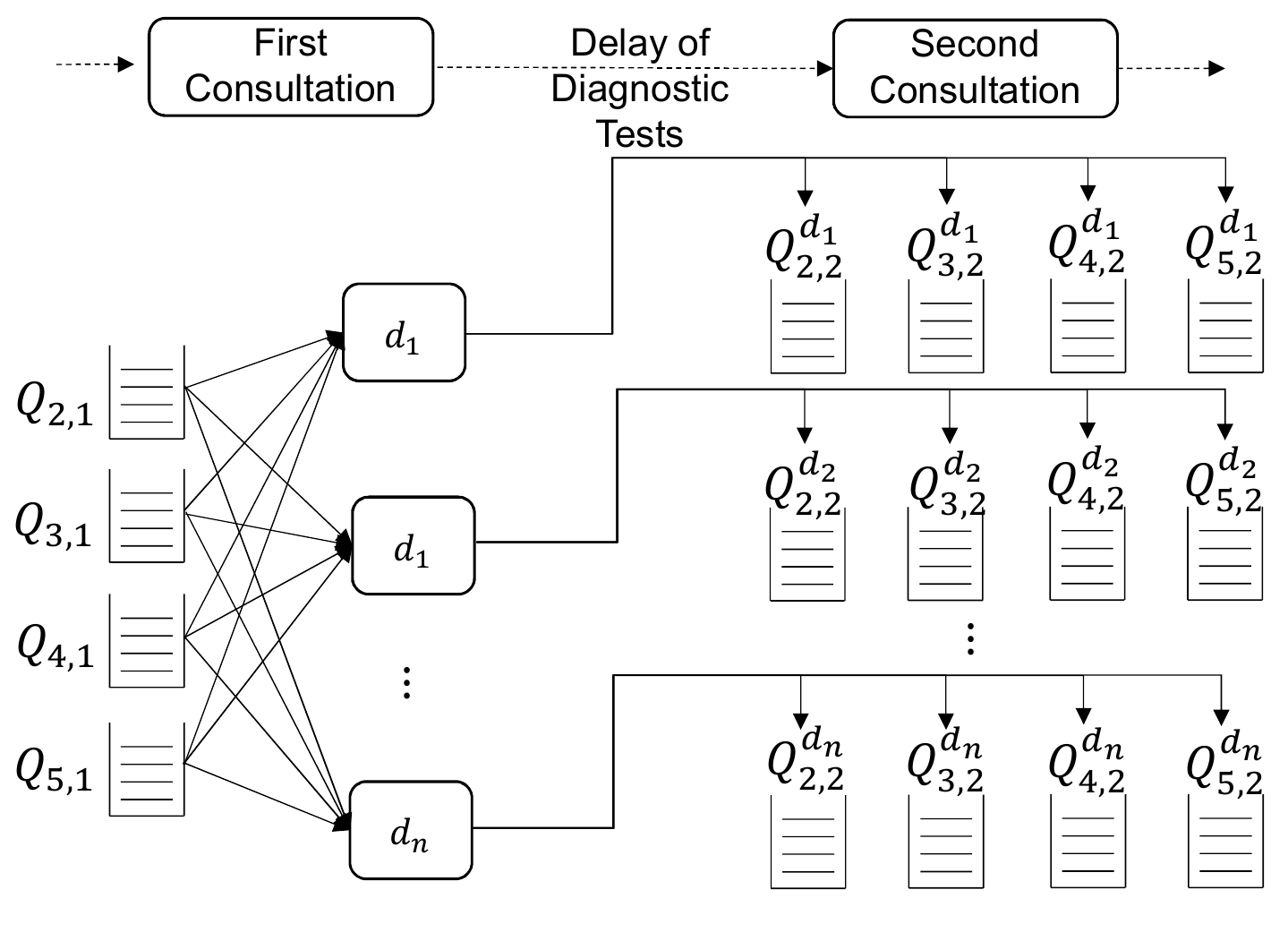}
    \caption{\ac{ED} Queue Structures Corresponding to Triage Grades and Doctors}
    \label{fig:QueueStructure}
    \end{center}
    
\end{figure}

Therefore, in this paper, patients waiting to be seen by a doctor are further split up in queues per triage category, i.e. $2$-$5$, by the type of consultation, i.e. first or second, and in case of a second consultation by the doctor they were seen during the first consultation (due to the ``same-doctor-policy''), similar to \cite{Cildoz2019}. The resulting queuing system of $4+4|D|$ queues is illustrated by figure \ref{fig:QueueStructure}.

In the setting illustrated by Figure \ref{fig:QueueStructure}, and if we assume that $d \in D$ is an idle doctor and $S_{ED}^t$ fully describes the state of the \ac{ED} at time $t$, a patient selection policy for $d$ may be described by a function

\begin{equation} \label{eq::PatientSelection}
    f_{ps}^d: S_{ED}^t \rightarrow \{Q_{2,1}, Q_{3,1}, Q_{4,1}, Q_{5,1}, Q_{2,2}^d, Q_{3,2}^d, Q_{4,2}^d, Q_{5,2}^d\}.
\end{equation}

Note that the above definition of function $f_{ps}^d$ implies some limitations and constraints. First, defining $f_{ps}^d$ per doctor implies that idle doctors are dispatched one at a time, e.g. by a possible ordering of idle doctors. In particular, we assume that doctors are dispatched in the order by their idle times, i.e. the doctor being idle the longest is considered first. This assumption prevents the use of policies based on more complex resource selection mechanisms, e.g. team selection \cite{Holm2009} or dynamic work load calculation \cite{Kang2016}. However, such complex resource selection policies are beyond the scope of this study and may be included in future research.

Second, the definition of $f_{ps}^d$ implies that patients within a specific queue are chosen with respect to the FIFO principle, i.e. within one queue always the patient waiting longest is seen first. This means that a selection policy may let patients with a lower priority ``overtake'' a patient with a higher priority, as they are not in the same queue, but within a single queue ``overtaking'' is not allowed. Note that due to stochastic registration and triage durations, the order of patients in a queue $Q_{j,k}$ for $j \in J, k \in \{ 1,2 \}$ may not be the same as the order of arrival of those patients, i.e. implied by $a_i$ for $i \in Q_{j,k}$. Analogously, the order of patients of triage category $j$ seen by doctor $d$ may be different to the order in which they were seen for the first consultation due to stochastic diagnostic times. In particular, the order of patients in $Q_{j,1}$ are implied by $t_{sw_3}^i$ and by $t_{sw_4}^i$ for patients in $Q_{j,2}^d$ for $d \in D$.

In the following, two established versions of function $f_{ps}^d$ are introduced, which serve as a benchmark for the proposed \ac{ML} based patient selection method.

\subsubsection{Basic Queuing Principles}

Basic queuing principles, or ``pure priority based orderings'' \cite{Cildoz2019}, define $f_{ps}^d$ in a way that queues are always considered in a fixed (priority based) ordering, and the first non-empty queue is selected when dispatching an idle doctor. Hence, at time $t$, a patient in $Q_{j_1,1}$ will always be seen before a patient in $Q_{j_2,1}$ if $j_1<j_2$, $|Q_{j_1,1}| >0$ and $|Q_{j_2,1}| >0$ (analogously for second consultation queues). Different versions of basic queuing principle are a result of ordering first and second consultation queues, often referred to as ``prioritizing old vs. new patients'' or vice-versa, see for example \cite{Chonde2013,Facchin2010, Ghanes2014,Saghafian2012,Tan2012,Vrugt2016,Yang2016}.

\cite{Cildoz2019} identify 4 different pure priority based strategies based on three different triage categories, where one is designed to reflect the actual queuing policy at the \ac{ED} their study is built on. Based on their definitions we propose 4 orderings of queues $Q_{j,k}$ for $j \in J, k \in \{ 1,2 \}$ that are summarized in Table \ref{tab::BasicQueuing}.

\begin{table}[ht!]
	\begin{center}
		\begin{tabular}{l l l l l l l l l}
			\hline
			 Policy & \multicolumn{8}{c}{Ordering} \\
			 \hline
			   & 1st & 2nd & 3rd & 4th & 5th & 6th & 7th & 8th \\
			   \hline
			   $QP1$ & $Q_{2,1}$ & $Q_{3,1}$ & $Q_{4,1}$ & $Q_{5,1}$ & $Q_{2,2}$ & $Q_{3,2}$ & $Q_{4,2}$ & $Q_{5,2}$ \\
			   $QP2$ & $Q_{2,1}$ & $Q_{2,2}$ & $Q_{3,1}$ & $Q_{3,2}$ & $Q_{4,1}$ & $Q_{4,2}$ & $Q_{5,1}$ & $Q_{5,2}$ \\
			   $QP3$ & $Q_{2,2}$ & $Q_{3,2}$ & $Q_{4,2}$ & $Q_{5,2}$ & $Q_{2,1}$ & $Q_{3,1}$ & $Q_{4,1}$ & $Q_{5,1}$ \\
			   $QP4$ & $Q_{2,2}$ & $Q_{2,1}$ & $Q_{3,2}$ & $Q_{3,1}$ & $Q_{4,2}$ & $Q_{4,1}$ & $Q_{5,2}$ & $Q_{5,1}$ \\

			 \hline
		\end{tabular}
		\caption{Basic Queuing Policies Defined by Their Queue Ordering}
		\label{tab::BasicQueuing}
	\end{center}
\end{table}
  
Policies $QP1$ and $QP3$ order queues $Q_{j,k}$ first with respect to $k$, i.e. type of consultation, and within those categories with respect to triage grades $j$. Hence, $QP1$ will always select ``new'' patients over old patients, and $QP3$ vice-versa. On the other hand, $QP2$ and $QP4$ first order queues by triage grade $j$ followed by the type of consultation $k$. Within a triage category $QP2$ prioritizes ``new'' over ``old'' patients, and $QP4$ ``old'' over ``new''. Note that other orderings than $QP1-QP4$ would be feasible, but arguably lack practicability. 
  
\subsubsection{\acl{APQ}}

The \ac{APQ} policy generalizes pure priority queuing by relaxing the constraint that the selection of patients of $Q_{j,1}$ with $j \in J$ is strictly done with respect to triage grades (and analogously for selecting patients of $Q_{j,2}^d$ with $j \in J, d \in D$), see for example \cite{Cildoz2019}. In particular, patients accumulate priority points over time at a predefined rate that is constant per patient class. When a doctor becomes idle, the patient with the most priority points is selected for consultation. Thereby, patients may ``overtake'' patients with higher priorities, depending on waiting times and accumulation rates. Analogously to \cite{Cildoz2019}, we define patient classes over triage grades and type of consultation. Hence, there is exactly one patient class for each type of queue $Q_{j,1}$ with $j \in J$ and one patient class for each set of queues $\{Q_{j,2}^d|d\in D \}$  with $j \in J$. Let $\beta_{j,k}$ be the corresponding accumulation rates. The accumulated priority points for a patient $i$ with triage grade $j$ waiting for consultation type $k$ at time $t$, is then given by the linear function $\beta_{j,k}(t-t^i_{a})$. Further, let $Q^d=\{Q_{j,1}|j \in J\} \cup \{Q^d_{j,2}|j \in J\}$ be the set of all queues doctor $d$ may select patients from, $\hat{Q}^d$ the non-empty queues for doctor $d$ (with at least one patient waiting), $\beta_q$ the corresponding weight and $p_q^1$ the first patient of a queue $q \in \hat{Q}^d$. The selection function $f_{ps}^d$ at time $t$ for doctor $d$ in case of \ac{APQ} patient selection can be written as

\begin{equation}
    f_{APQ}^d =
\begin{cases}
\argmax_{q \in \hat{Q}^d } \left( \beta_q (t - t_{a}^{p_q^1}) \right), \text{    if  } |\hat{Q}^d| > 0\\
None,  \text{    otherwise.}
\end{cases}
\end{equation}

Note that the above definition, still satisfies constraints that patients of a particular queue $q \in Q^d$ are seen in the order they join the queue.

The used procedure to optimize weights $\beta$ and corresponding results are outlined in the Appendix of the paper.

\section{\ac{ML} Based Patient Selection} \label{sec::MLPatientSelection}

In its essence, \ac{APQ} patient selection aims to balance the impact of patients of different triage grades on a chosen objective function, thereby improving the overall performance of the \ac{ED} with respect to that measure. Therefore, additional information on the system's state, i.e. patients' time in the system, are used within a linear function as a basis for decision making. Naturally, the question arises, if such a method could be improved by considering an extended information base and using more complex functions than linear equations.

In an ideal, but not realistic, setting, one would not only have perfect information on the current state of the \ac{ED}, but also information on future events in the \ac{ED}. If such information was available, patient selection would essentially become a question of scheduling tasks to resources, e.g. doctors. In that case, a deterministic optimization algorithm, exact or heuristic, could be used to optimize $\hat{f}_{ED}$, and make optimal (or at least close to optimal) patient selection decisions at any time. However, as mentioned before, this would require that all future arriving patients, all activity durations and requirements of diagnostic tests are known at the time of planning, i.e. when scheduling patient requests. 

Obviously, such information will never be available at the time of decision making in a real-world setting with stochastic behavior. However, having a model representing the real system allows for the generation of a (theoretically infinite) set of instances of the studied \ac{ED}. For those instances, one is able to use a deterministic optimization algorithm to compute optimal (or close to optimal) patient selection decisions with respect to $\hat{f}_{ED}$ and thereby generate an information basis for enhanced decision making.

\begin{figure}[ht!]
    \begin{center}
    \includegraphics[width=0.8\textwidth]{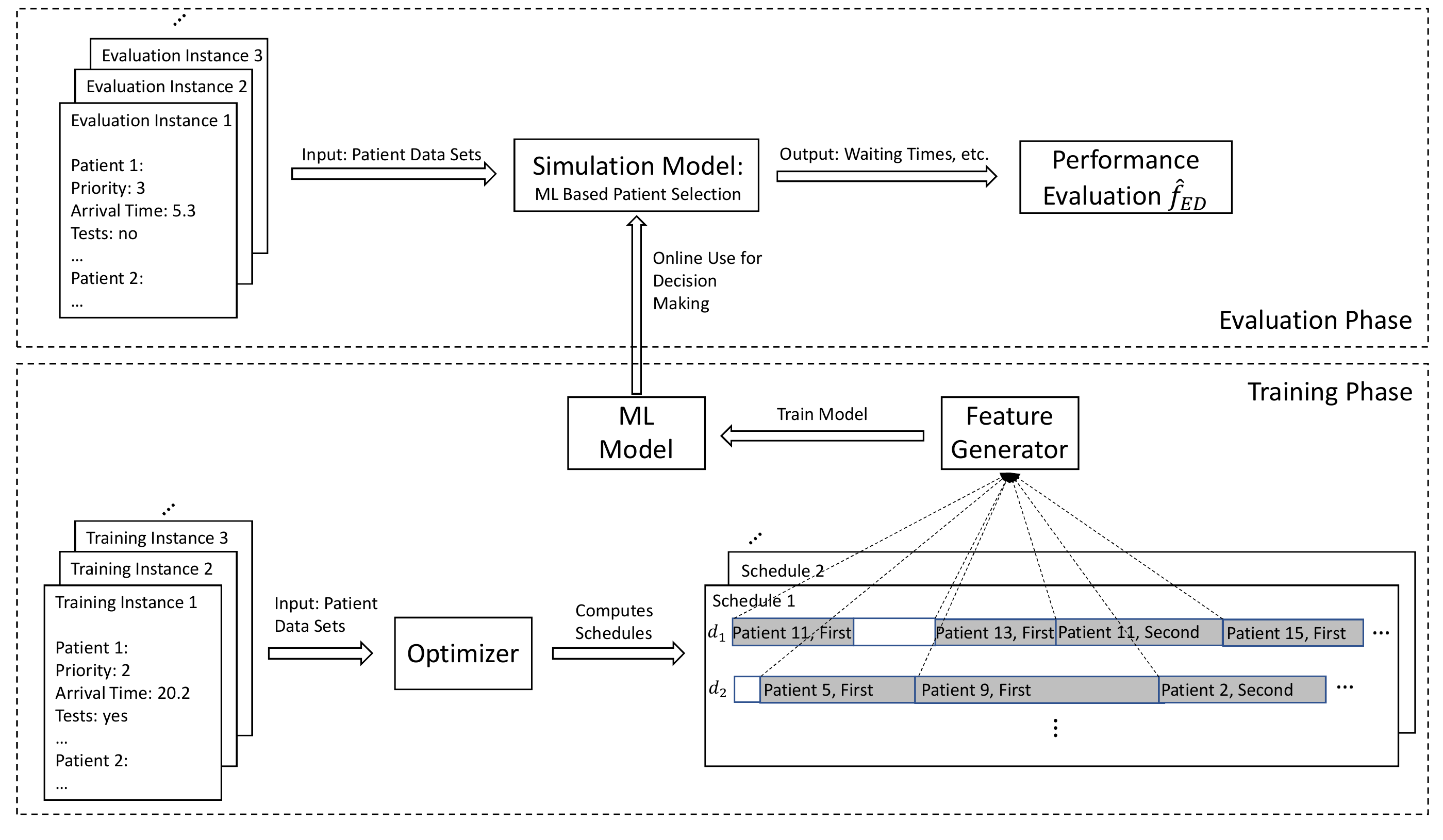}
    \caption{General Principle and Architecture of the Design and Evaluation of the \ac{ML} Based Patient Selection Policy}
    \label{fig::MLPrinciple}
    \end{center}
    
\end{figure}

The proposed \ac{ML} based patient selection policy learns from optimal decisions made by a deterministic optimizer during a training phase and transforms the gained ``knowledge'' into a \ac{ML} model. This model can then be used for patient selection for future unseen instances. Thereby, it extends the considered system state of \ac{APQ} by any feature set describing the \ac{ED} at a given point in time and replaces the linear function of \ac{APQ} by a more complex and non-linear function, i.e. the \ac{ML} model. The main principle of the approach is outlined by figure \ref{fig::MLPrinciple}

During a training phase a heuristic optimization algorithm is used to solve a set of randomly generated training instances. The generation of instances uses the same stochastic processes and probability distributions that are embedded in the simulation model and will be described in detail in section \ref{sec::data}. Hence, for a chosen period of time, a set of patients with associated parameters, i.e. arrival times, triage grades activity durations, and diagnostic test requirements, are sampled per instance from those distributions and processes. Since we are assuming time-of-the-day dependent arrival rates of patients (see section \ref{sec::data}), the smallest reasonable period of time covered by one instance is 24 hours. Instances covering more than 24 hours increase the complexity of the scheduling problems to solve, without adding much value to the gained information. Hence, generated instances include patients arriving during a 24 hour period starting at midnight. It is important to note that resulting schedules may stretch over a longer period of time, as all activities of patients arriving within that 24 hour period are scheduled. This results in a slight inaccuracy for the last activities to schedule, as newly arriving patients outside the time window of 24 hours are not considered. However, as \ac{ED} are typically less frequently visited during night hours (as will be apparent by the chosen patient arrival pattern), this small bias should be negligible.

The optimizer treats the generated instances as a deterministic input to a scheduling problem. The scheduling of registration and triage tasks is done with respect to the FIFO principle, but the assignment of consultation tasks to doctors is subjected to optimization. The same set of constraints as for other queuing policies, i.e. no overtaking within queues and the same-doctor-policy is considered by the optimizer. The result of this process are schedules per doctor, i.e. at what point in time which patient is seen, for each generated training instance.

The feature generator transforms the optimized schedules to a set of features. Each point in time where a patient starts a consultation activity with a doctor corresponds to a decision made in the abstracted system. Therefore, the feature generator iterates over each of those time steps, stores information of the state of the system at the specific point in time as a sample and the ``optimal'' decision as a response. Thereby, the decision is represented by the queue identifier of the queue the patient is selected from. Note that, for defining responses, we do not distinguish between queues per doctors for second consultations. Hence, there are eight possible responses $QR = \{ Q_{2,1} , Q_{3,1} , Q_{4,1} , Q_{5,1} , Q_{2,2} , Q_{3,2} , Q_{4,2} , Q_{5,2} \}$. 

The resulting sample and response training set is then used to fit a \ac{ML} classifier model with eight possible classes, i.e. $QR$.

In the evaluation phase, the \ac{ML} classifier is used within the simulation model of the \ac{ED} to select patients from queues whenever a doctor becomes idle and there are patients either waiting for the first consultation or the second consultation with that particular doctor. The simulation model is then evaluated on a set of evaluation instances that are generated with respect to the same distributions and processes as the training instances. Note that those steps are not only dependent on the input data, e.g. stochastic components used, but also on the used weights of the objective function $\hat{f}_{ED}$. This requires the optimization of training instance and fitting of the \ac{ML} for each parametrization of $\hat{f}_{ED}$ with different weights.

In the following, the most important components of the proposed \ac{ML} based patient selection method are described in detail individually.

\subsection{Optimizing \ac{ED} instances}

The training base for the proposed \ac{ML} patient selection method is built on optimal, or near optimal, schedules for one day instances of the \ac{ED}. The underlying optimization problem is basically a job-shop scheduling problem of assigning consultation tasks (jobs) to doctors (machines or resources). Exact methods to solve such problems often rely on column generation within a branch-and-price or branch-and-price-and-cut framework. However, to use column generation, (or decomposition in general) effectively, it is usually required that schedules  can be independently computed and priced. Unfortunately, this is not the case for the underlying optimization problem of scheduling patient consultations to doctors. 

Re-ordering of jobs within a schedule of a single doctor may make schedules of other doctors infeasible, as patients within one queue need to be selected exactly in that order. The same observation holds for removing patients from a doctor's schedule and inserting it in another one's schedule. In particular, the earliest possible time a patient may be seen by a doctor may depend on the schedule of other doctors. This inter-dependency limits the usage of exact methods.

However, as results will show, even heuristically computed near-optimal solutions are effective as a training base for the \ac{ML} approach. Thus, we use a Simulated Annealing-based adaptive neighborhood search algorithm to heuristically optimize one-day instances of the \ac{ED}. Further, to stabilize the search, the Simulated Annealing method is embedded in a (simplified) race algorithm.

In the following, the solution decoding and encoding, the neighborhood definitions, adaptive mechanisms and the final search heuristic are described in detail.

\subsubsection{Solution Coding and Decoding} \label{sec::Decode}

The majority of meta-heuristics used for optimization (as Simulated Annealing) require a solution representation that can easily be decoded to an actual solution and vice-versa. A key requirement of such a solution representation is to allow the definition of effective operators, such as neighborhood operators. Initially, we experimented with solution representations that were based on actual schedules per doctors. However, the intra-queue constraints of the underlying problem made the definition of operators cumbersome, and led to an inferior performance of the search algorithm. 

Hence, the presented heuristic is based on a simpler representation of solutions. As the set of patients and their treatment requirements are treated as an input and are known, the number of patients that will eventually be in a queue $q \in QR = \{ Q_{2,1} , Q_{3,1} , Q_{4,1} , Q_{5,1} , Q_{2,2} , Q_{3,2} , Q_{4,2} , Q_{5,2} \}$ is also known upon execution of the algorithm. Therefore, we encode a solution as the sequence of queues, where the number of occurrences of a queue identifier $q \in QR$ matches exactly the number of consultation requirements of patients of the corresponding triage grade and consultation type. For example, if during the day in total 20 triage grade 4 patients arrive at the ED, out of which 15 require diagnostic tests and a second consultation, any sequence representing a solution to that instance contains exactly 20 $Q_{4,1}$ and 15 $Q_{4,2}$ entries. 

It is important to note that such a sequence is only defining the order of queues by which patients get selected and is not explicitly defining the order by which patients are selected. Further, queues for the second consultation are not considered by doctors in the sequence, i.e. there are no distinct identifiers for each doctor.

To decode such a sequence to an actual solution, i.e. a schedule, the simulation model described in Section \ref{sec::EDModel} is used. But instead of using a pure priority or \ac{APQ} policy for dispatching idle doctors, a sequence based policy is used. Therefore, the sequence of queues is passed to the simulation model upon initialization and the model is started. Whenever a doctor $d$ becomes idle, the patient to be seen is determined in the following way. The sequence is searched for the first entry where the corresponding queue is not empty at the current time. For entries representing queues for the second consultation, only the queue for doctor $d$ is considered. For example, if the first entry in the sequence is $Q_{3,2}$ but queue $Q_{3,2}^d$ is empty, the search moves to the next entry of the sequence. When the first non-empty queue is found, the first patient in that particular queue, i.e. the patient waiting longest, is seen by doctor $d$. Finally, the corresponding queue identifier is removed from the sequence and the simulation model continues.

\begin{figure}[ht!]
    \begin{center}
    \includegraphics[width=0.8\textwidth]{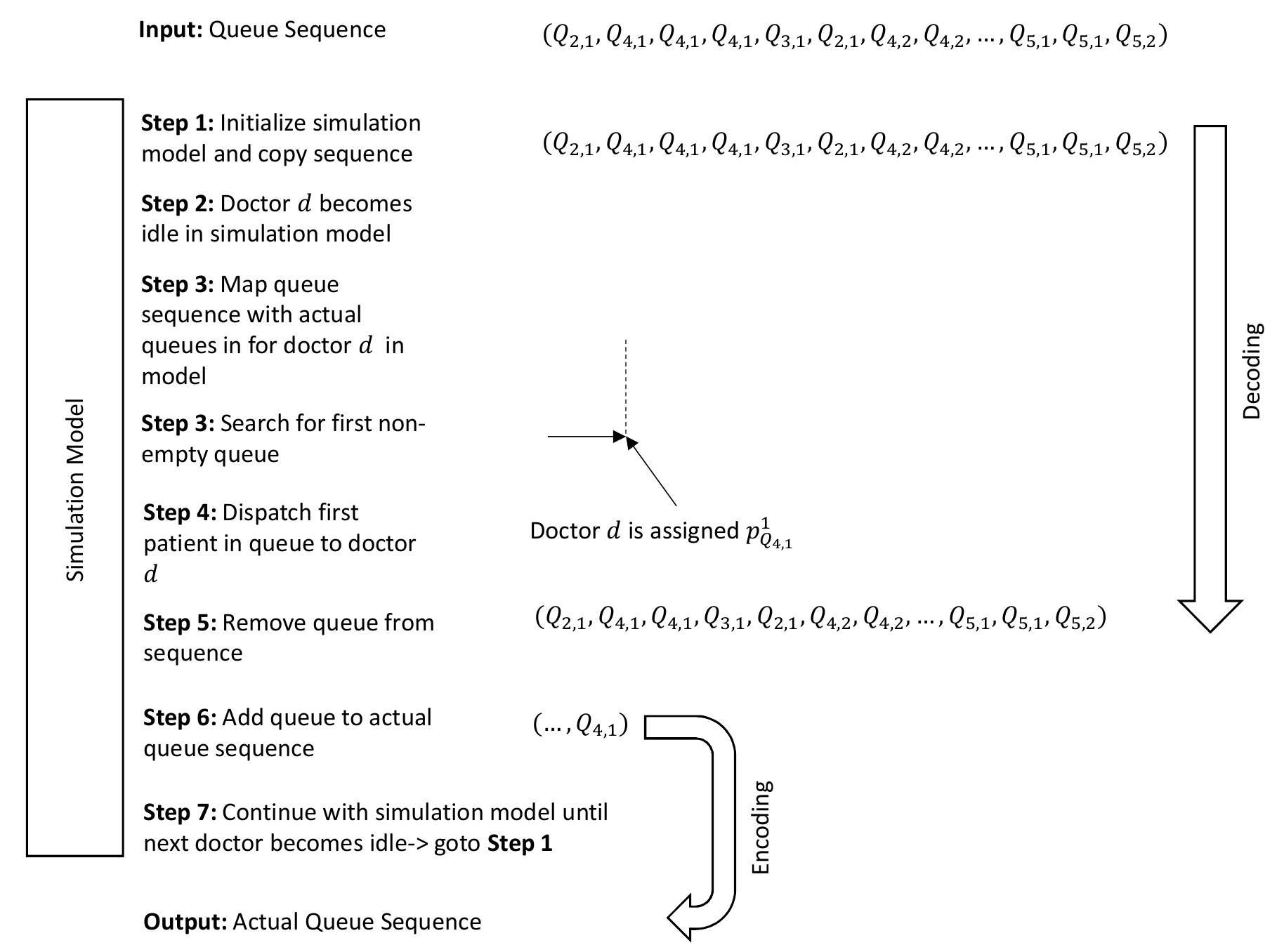}
    \caption{Simulation Based Solution Decoding and Encoding}
    \label{fig::SolDecoding}
    \end{center}
    
\end{figure}

Since sequences are subjected to neighborhood operators, it regularly occurs that the actual queue a patient is selected from is not the first queue in the sequence. In other words, the sequence of queues by which patients are actually selected might differ from the sequence passed to the simulation model. To improve the search, whenever a solution representation is decoded, the actual queue sequence is stored along with the solution itself, i.e. the schedule, and returned by the simulation model upon termination. The overall principle is illustrated by Figure \ref{fig::SolDecoding}.

\subsubsection{Neighborhood Definitions}

Based on the solution representation and decoding principle described in the previous section, we will define in total four different neighborhood operators. A common principle for sequence or permutation based representations, is to define neighborhood operators based on swapping of elements, or removal and insertion of elements, in the sequence or permutation.

Basically, the proposed operators follow that principle. However, as swapping for example two $Q_{4,1}$ identifiers will not change the sequence, we define operators based on positions in the sequence where two consecutive identifiers differ. In the following, such positions are referred to as ``change points''.

The first operator, referred to as the block neighborhood, randomly picks a change point in the sequence and swaps a chosen number of queue identifiers before the change point, i.e. a block, with the same sized block after the change point. The size of blocks is randomly picked from an interval $ \left[1,b \right]$, where $b$ is a parameter that is adapted along the search.

The other operators are based on a removal and insertion principle. First, a change point is randomly selected. Second, a block size is randomly selected from $ \left[1,b \right]$ and queue identifiers in the block before and after the change point are removed. Finally, a new sub-sequence is formed by the removed identifiers and this sequence is inserted in the original sequence at the position of removal. 

In general, the sub-sequence is constructed randomly, but three different versions are considered:
\begin{enumerate}
    \item The ``random insertion operator'' inserts the removed identifiers in a purely random fashion.
    \item The ``objective insertion operator'' assigns weights to the identifiers that are based on the objective value of the current solution. In particular, each identifier is associated with an actual patient $i$ in the decoded solution. The contribution of that particular patient to the overall objective $\hat{f}_{ED}$, i.e. the term corresponding to patient $i$ in equation \ref{eq::EmpObj}, is used as a weight. The sum of weights of patients in the removed block is normalized to 1 and probabilities are assigned correspondingly. Identifiers are then iteratively picked and added to the new sub-sequence with respect to those probabilities.
    \item The ``queue insertion operator'' works similar as the objective based operator, but instead of using the contribution to the objective value as weights for queue identifiers, queue lengths are used as weights. Therefore, the state of queues in the \ac{ED} is checked after dispatching the last patient before the removed block in the decoded solution of the current sequence. The number of patients in corresponding queues at that point in time in the solution are then used as weights for removed queue identifiers. Note that queues for different doctors are aggregated to a single weight, i.e. summed up.
\end{enumerate}

\begin{figure}[ht!]
    \begin{center}
    \includegraphics[width=0.8\textwidth]{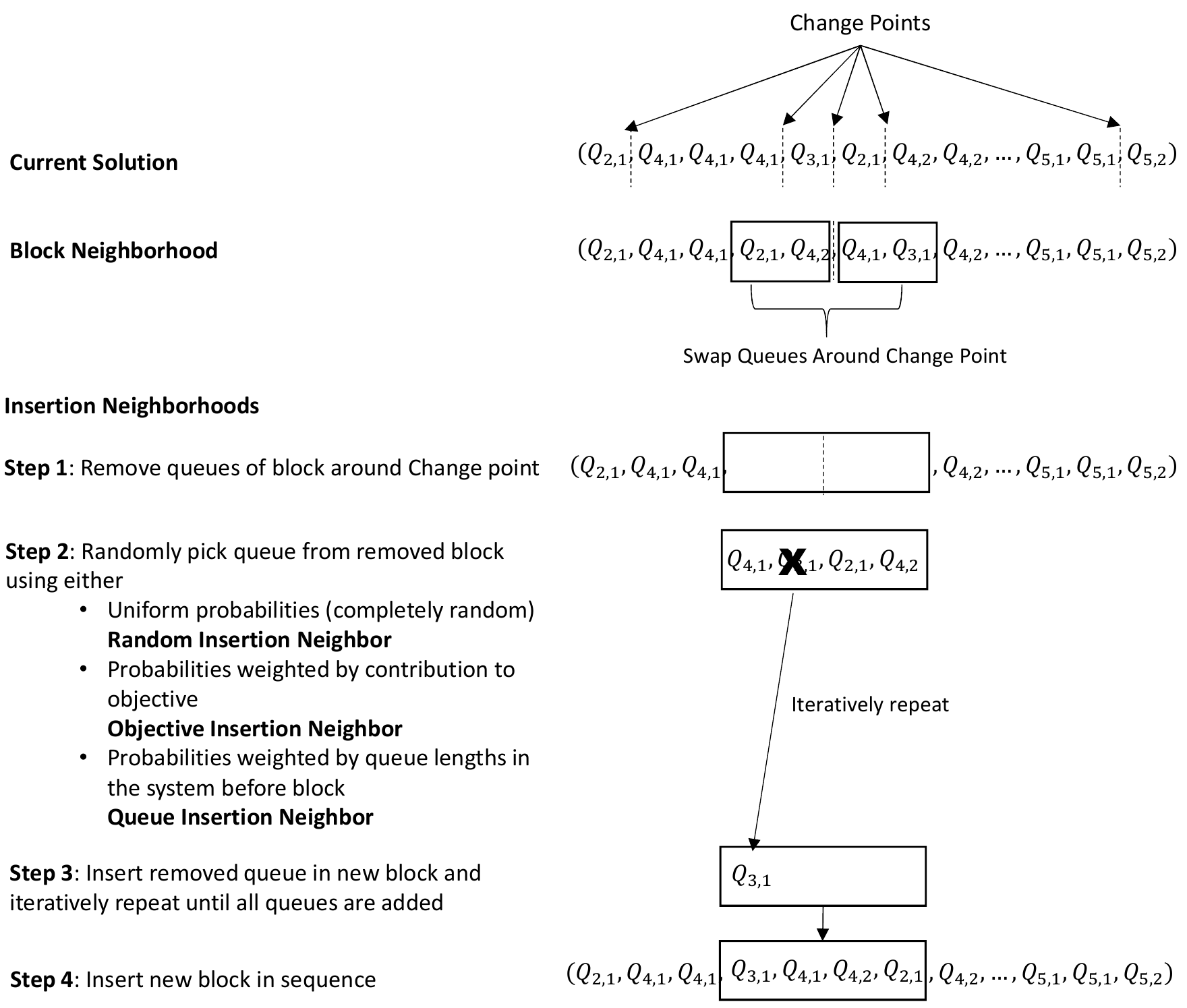}
    \caption{Definition of Neighborhood Operators}
    \label{fig::NeighborOperator}
    \end{center}
    
\end{figure}

The general principle of change points and the four neighborhood operators is outlined by Figure \ref{fig::NeighborOperator}.

\subsubsection{Competitive Simulated Annealing Heuristic}

The heuristic to optimize \ac{ED} instances proposed in this paper is based on the Simulated Annealing meta-heuristic. In its simplest form, Simulated Annealing iteratively updates a current solution by randomly picking a neighborhood solution (e.g. via randomly applying a neighborhood operator) and comparing its objective with the objective of the current solution. If the objective is improving, the move to the new solutions is accepted, i.e. the current solution is updated. If the objective is non-improving, the move is only accepted with a certain probability that is dependent on the distance between objectives and a so-called temperature. This temperature is assumed to decrease over iterations, i.e. making it less likely for worse solutions to be accepted. 

Several adaptions and extensions to that base principle have been proposed to solve real-world problems. Two commonly integrated adaptions, that are also used in the presented version, are adaptive mechanisms and diversification moves.

Adaptive mechanisms are usually used to pick operators from a set of operators at a specific iteration along the search with respect to their performance at previous iterations, see for example \cite{Azi2014}. Assuming that $h$ such operators are available, each with a weight $w_k, k=1,\ldots,h$, the probability of choosing operator $k$ is given by

\begin{equation} \label{eq::AdaptiveWeights}
    \frac{w_k}{\sum_{i=1}^h w_i}. 
\end{equation}

Weights $w_k$ are initialized uniformly and updated after a consecutive number of iterations, referred to as a segment. After each segment, weights are updated by

\begin{equation} \label{eq::AdaptiveUpdatesWeights}
    w_k^{sg} = \gamma w_k^{sg-1} + (1- \gamma) \pi_k^{sg-1},
\end{equation}

where $w_k^{sg-1}$ are the weights of the previous segment, $\gamma$ is a parameter, and $\pi_k^{sg-1}$ is a performance score of operator $k$ during the previous segment, see for example \cite{Azi2014}. At the start of each segment $\pi_k^{sg-1}$ are set to $0$ and updated each iteration by the following rule

\begin{equation} \label{eq::AdaptiveUpdatesScore}
    \pi_k^{t+1} = \pi_k^t + 
    \begin{cases}
    \sigma_1 \text{  if a new best solution has been produced,} \\
    \sigma_2 \text{  if the new solution is better than the current solution,}\\
    \sigma_3 \text{  if the new solution is worse than the current solution, but has been accepted,}
    \end{cases}
\end{equation}

where $\sigma_1$, $\sigma_2$, and $\sigma_3$ are parameters to chose.

We use this adaptive mechanism to select a neighborhood operator from the following operators: block neighborhood, random insertion neighborhood and objective insertion neighborhood. Based on experimentation, the segment size is chosen as 5000 iterations, $\gamma$ is chosen as $0.95$ and sigma weights are chosen as $0.5$, $2$, and $0.25$.

However, we use the same principle not only for selecting operators, but also for randomly selecting a change point in the current sequence. Therefore, the positions in a sequence reflected by change points are kept in buckets of size 10, e.g. change points at position 22 and 27 would both be in the same bucket. The number of buckets is limited by the length of the sequence divided by 10 plus 1. Weights are kept and updated for each bucket in the same way as weights for operators are maintained. When selecting a change point, first a bucket is chosen with respect to current weights, and second a change point is randomly chosen within the bucket with respect to uniform probabilities. Parameter $\gamma$ is set to $0.99$ and parameter $\sigma_2$ is set to $1$, where all other parameters are the same as for the selection of neighborhood operators.

The queue insertion neighborhood operator is used within diversification steps. In particular, whenever the search fails to improve the overall best solution for consecutive $d_{max}$ iterations, a diversification move is performed. Such moves include the application of the queue based insertion neighborhood operator with block size $b_{max}$ and resetting weights for operators and change point buckets to uniformly distributed weights.

As a starting solution, the best solution of the set of solutions resulting from applying the pure queuing principles $QP1$-$QP4$ is chosen. The resulting procedure is described by algorithm \ref{alg::SimulatedAnnealing}. Therefore, let $Decode(seq)$ be a simulation based method that decodes a sequence of queues based on the principle described in section \ref{sec::Decode} and returns the corresponding objective value and actual queue sequence.

\begin{algorithm}
\caption{Adaptive Simulated Annealing}
\label{alg::SimulatedAnnealing}
\begin{algorithmic}[1]
\STATE{\textbf{Input:} Instance $I$, $init$, $iter_{max}$}
\STATE{\textbf{Output:} Optimized Schedule}
\IF{$init$}
\STATE{\textbf{Set} $iter=0$, $b=b_{max}$, $lastImprov = 0$, $T=2$}
\STATE{\textbf{Initialize} Weights for NB Operators and Change Point Buckets}
\STATE \textbf{Compute} start solution $seq_{start}$ with objective $f_{start}$
\STATE \textbf{Set} $best_{seq} = seq_{start}, f_{best} = f_{start}, cur_{seq} = seq_{start}, f_{cur} = f_{start}$
\ELSE
\STATE \textbf{Restore} $iter, best_{seq}, f_{best}, cur_{seq}, f_{cur}, T, lastImprov$, NB and Change Point Weights and Scores from Last Run
\ENDIF
\WHILE{$iter < iter_{max}$}
\STATE \textbf{Set} $iter = iter +1$ and $T= \max(T \cdot 0.0.999975,0.3)$

\IF{$iter \mod 5000 = 0$}
\STATE \textbf{Update} Weights for NB and Change Points by (\ref{eq::AdaptiveUpdatesWeights})
\ENDIF
\STATE \textbf{Select} NB Operator with weights by (\ref{eq::AdaptiveWeights})
\STATE \textbf{Select} Change Point Bucket with weights by by (\ref{eq::AdaptiveWeights}) and Change Point in Bucket
\STATE \textbf{Choose} NB Size as $randInt(1,b)$
\STATE \textbf{Compute} New Sequence $new_{seq}$ with Selected NB and Size
\STATE \textbf{Set} $new_{seq},f_{new} = Decode(new_{seq})$
\IF{$f_{new} < f_{cur}$}
\STATE \textbf{Set} $f_{cur} = f_{new}, cur_{seq} = new_{seq}, b = \max(1,b-1)$
\IF{$f_{new}< f_{best}$}
\STATE \textbf{Set} $best_{seq} = new_{seq}, f_{best} = f_{new}, lastImprov = iter$
\ENDIF
\ELSIF{$rand(0,1) < \exp^{-(f_{new} - f_{cur})/T}$}
\STATE \textbf{Set} $f_{cur} = f_{new}, cur_{seq} = new_{seq}$
\ELSE
\STATE \textbf{Set} $b=\min(b+1, b_{max})$
\ENDIF
\STATE \textbf{Update} Scores $\pi$ for NB and Change Points with (\ref{eq::AdaptiveUpdatesScore})
\IF{$iter - lastImprov > d_{max}$}
\STATE \textbf{Set} $lastImprov = iter$, $b= b_{max}$, $T = \min(T\cdot2, 2)$
\STATE \textbf{Reset} Weights for NB and Change Points to Uniform Weights
\STATE \textbf{Set} Scores for NB and Change Points to $0$
\STATE \textbf{Compute} New Sequence $new_{seq}$ with Queue Based Insertion NB
\STATE \textbf{Set} $new_{seq},f_{new} = Decode(new_{seq})$
\STATE \textbf{Set} $f_{cur} = f_{new}, cur_{seq} = new_{seq}$
\ENDIF
\ENDWHILE
\STATE \textbf{Return} $best_{seq}, f_{best}$
\end{algorithmic}
\end{algorithm}

By its definitions, Simulated Annealing  in general as well as the proposed adaptive version outlined by Algorithm \ref{alg::SimulatedAnnealing} are stochastic algorithms. Hence, multiple runs on the same instance (with varying random seeds) may lead to different results. In order to stabilize the search, Algorithm \ref{alg::SimulatedAnnealing} is embedded in a race-like algorithm. In particular, the search of Algorithm \ref{alg::SimulatedAnnealing} is divided in three phases of iterations. Let $iter_{max}^1$, $iter_{max}^2$ and $iter_{max}^3 = iter_{max}$ denote the number of iterations of each phase. Initially, $n_s^1$ instances of Algorithm \ref{alg::SimulatedAnnealing} are initialized (with varying random seeds) and for each instance $iter_{max}^1$ iterations are performed. After completion of those $iter_{max}^1$ iterations, the current best objectives, $f_{best}$ of all instances are compared and the $n_s^2$ best instances are passed to the second phase. For those, the search is continued until $iter_{max}^2$ iterations are reached. Only the instances of the search with $f_{best}$ among the $n_s^3$ best objectives are kept and passed to the last phase. During that last phase the search is continued until the overall maximum number of iterations, $iter_{max}$, is reached. The final best solution found is chosen from those $n_s^3$ instances in the last phase. The parameters used in this paper are $(iter_{max}^1, iter_{max}^2,iter_{max}) = (25000,50000,20000)$ and $(n_s^1,n_s^2,n_s^3) = (10,3,1)$.

\subsection{\ac{ML} Model Integration}

Given that doctor $d \in D$ is idle at time $t$, the task of the \ac{ML} is to pick the patient to be seen next by doctor $d$. However, as overtaking within queues is not allowed, this task can be represented as picking a (non-empty) queue from $\{ Q_{2,1} , Q_{3,1} , Q_{4,1} , Q_{5,1} , Q^d_{2,2} , Q^d_{3,2} , Q^d_{4,2} , Q^d_{5,2} \}$. The first patient in the chosen queue is then selected for consultation (first or second) by the idle doctor.

Picking a queue from the reference set can also be seen as a classification task. In other words, for any possible situation in the \ac{ED}, i.e. any possible state, there exists at least one optimal selection, and the \ac{ML} aims to predict this optimal decision for a given state.

There exist several classifier models in \ac{ML} literature, such as, for example, decision trees, random forests, neural networks, logistic regression and support-vector-machine.

Experiments have shown that, apart from single decision trees, all of the above mentioned approaches lead to very similar results in terms of accuracy. In this paper a random forest approach is use. The exact definition of the feature set used for the predictor model is outlined in the Appendix of the paper.

Note that it is theoretically possible that a random forest model predicts, or selects, an empty queue for a given unseen state. Obviously, an empty queue is never part of a solution generated from the optimizer, and hence part of the training set of samples. However, due to techniques like feature sampling and bagging, during the generation of trees within the forest, it is very unlikely, but possible to occur. To avoid the inefficiency of picking an empty queue, the classes $\{ Q_{2,1} , Q_{3,1} , Q_{4,1} , Q_{5,1} , Q^d_{2,2} , Q^d_{3,2} , Q^d_{4,2} , Q^d_{5,2} \}$ are sorted by their prediction probabilities provided by the random forest model, and the first non-empty queue is picked. 

\section{Data Generation and Scenario Definition} \label{sec::DataAndScenarios}

In this section, we describe the parametrization of the model and the design of scenarios that are used for experimentation and evaluation of the proposed approach. In section \ref{sec::data} the ``base'' parametrization or scenario is described, whereas some of these inputs are altered in section \ref{sec::Scenarios} to define different scenarios. Presented inputs for the base scenario are not based on a real \ac{ED}, but are representative typical values that can be found in literature. Wherever possible, references to data sources are provided.

\subsection{\ac{ED} Data} \label{sec::data}

To paremeterize the model, three different sets of input measures have to be defined: patient related measures (arrival processes, triage grade distributions, etc.) activity durations, and resource availabilities. 

\begin{figure}[ht!]
    \begin{center}
    \includegraphics[width=10cm]{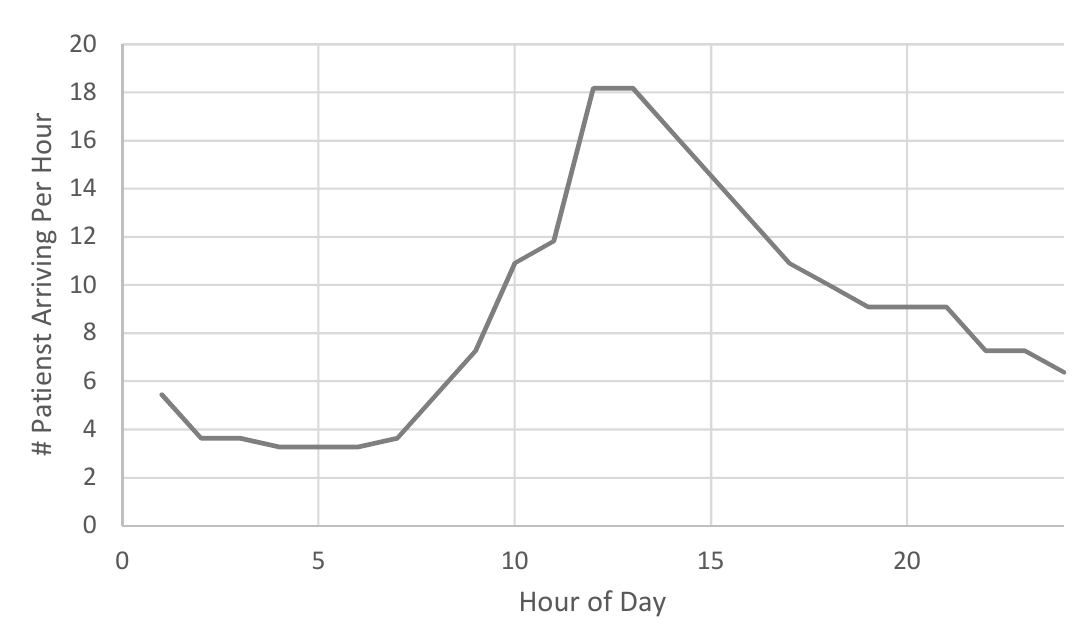}
    \caption{Arrival Stream of Patients}
    \label{fig::ArrivalStream}
    \end{center}
\end{figure}

Arrival streams of patients are most often modeled via a Poisson process with exponential inter-arrival times, see for example \cite{Yang2016,Facchin2010,Conelly2004,Vrugt2016,Saghafian2012}. Typical variations to identically distributed inter-arrival times are so called hour-of-day dependent arrival rates and/or day-of-week dependent arrival rates. In other words, it is typical that the number of patients arriving to an \ac{ED} is dependent on the time of the day (see for example \cite{Gunal2006,Carmen2015,Ashour2013,Chonde2013,Hoot2008b,Kang2016,ZEINALI2015123}) and sometimes also the weekday (\cite{Ghanes2014,Lim2013,Holm2009,Uriarte2017}). However, it is interesting to see that the distribution of arrival rates over the time of the day shows similar characteristics over various published \ac{ED} models, see \cite{Furian2018}, with a peak around noon and less patients arriving during night hours. Hence, we used the slightly adapted shape of the arrival rate curve proposed by \cite{Furian2019}, as shown by Figure \ref{fig::ArrivalStream}.

The distribution over triage grades of patients is chosen in the same way as proposed by \cite{Ferrand2018}, with the slight adaption that the probability mass of category 1 patients is shifted to category 2 patients. The probabilities of patients requiring a diagnostic test, and consequently a second consultation with a doctor, as well as the arrival modes are obtained from \cite{Gunal2006}. Note that the arrival mode has no influence on the processes of patients with a triage category other than 2. Target times and shares are chosen in the same way as by \cite{Cildoz2019}. The importance of patient categories $j \in J$ is based on the values used by \cite{Cildoz2019}, where for $j=2$ we defined the importance to be 3. All parameters and chosen importance of patient categories $pw_j$ are summarized by table \ref{tab::PatientDistributions}.

\begin{table}[ht!]

\begin{center}
    \begin{tabular}{l l l l l }
    \hline
    Triage Grade & 2 & 3 & 4 & 5 \\
    \hline
    Probability & 6.5\% & 20\% & 51.5\% & 22\% \\
    Probability Diagnostic Test & 74\% & 62\% & 53\% & 31\% \\
    Probability Ambulance Arrival & 60\% & - & - & - \\
    Patient Importance $pw_j$ & 3 & 2 & 1.5 & 1 \\
    $t^j_{ttd}$ & 15 & 30 & 60 & 120 \\
    $ts^j_{ttd}$ & 95\% & 90\% & 83\% & 80\% \\
    \hline
    \end{tabular}
    \caption{Probabilities, Importance, Target Time and Shares of Patient Classes}
    \label{tab::PatientDistributions}
\end{center}
\end{table}

The probability distributions for the duration of activities are summarized by table \ref{tab::ActivityDurations}. For register and triage activities we assumed that the duration follows a triangular distribution with associated parameters. The duration of registering is independent of the triage grade (which is also not known at the time of register) and parameters are chosen in the same way as in \cite{Furian2019}. The duration of triage activities is chosen to be slightly shorter than that proposed by \cite{Ferrand2018}, as this led to more realistic shares of patients meeting the \ac{TTD} targets.

Similar to \cite{Cildoz2019}, the duration of first and second consultation activities are assumed to be Log-Normally distributed. Thereby, the $\sigma$-parameter is chosen exactly as by \cite{Cildoz2019} for all consultation activities. The $\mu$ parameter was used to achieve the desired expected value of the specific duration. The first consultation task is assumed to have an expected duration of $17$ minutes for category 2 patients, as also proposed by \cite{Ferrand2018}, $20$ minutes for category 3 patients, as \cite{Cildoz2019}, $18$ minutes for category 4 patients (as a mix from $16$ minutes by \cite{Cildoz2019} and $20$ minutes of \cite{Ferrand2018}) and $15$ minutes for category 5 patients (as a mix from $13$ minutes by \cite{Cildoz2019}, $22$ minutes of \cite{Ferrand2018} and $10$ minutes of \cite{Furian2019}).

For the second consultation, the same values as reported in \cite{Cildoz2019} were used for category 3, 4 and 5 patients. A $12$ minute expected duration is assumed for category 2 patients (those are not included in the study of \cite{Cildoz2019}).

The delay of diagnostic tests (including possible waiting times) is modelled with a triangular distribution of mean $30$ minutes and $\pm15$ minutes as lower and upper limits.

\begin{table}[]
\begin{center}
    \begin{tabular}{l cccc}
    \hline
    Triage Grade        & 2 & 3 & 4 & 5 \\
    \hline
    Register            & - & $T(3,4,5)$ & $T(3,4,5)$ & $T(3,4,5)$  \\
    Triage              & $T(2,3,4)$ & $T(4,6,8)$  &  $T(3,5,7)$ & $T(3,5,7)$   \\
    First Consultation  & $LogN(2.73,0.45)$  &  $LogN(2.89,0.45)$ & $LogN(2.79,0.45)$  & $LogN(2.60,0.45)$  \\
    First Consultation $E$  & 17  &  20 & 18  & 15  \\
    Second Consultation & $LogN(2.38,0.45)$  &  $LogN(2.29,0.45)$ & $LogN(2.10,0.45)$  & $LogN(1.84,0.45)$  \\
    Second Consultation $E$  & 12  &  11 & 9  & 7  \\
    Diagnostic Tests    & $T(15,30,45)$ & $T(15,30,45)$  &  $T(15,30,45)$ & $T(15,30,45)$   \\
    \hline
    \end{tabular}
    \caption{Stochastic Distributions of Activity Durations: $T(l,m,u)$ Triangular Distributions (lower, mode, upper), $LogN(\mu,\sigma)$ Log-Normal Distribution with Expected Value $E$ in minutes}
    \label{tab::ActivityDurations}
\end{center}
\end{table}

We did not include shift-models in the \ac{ED} models, and assumed that 4 doctors, 2 triage nurses and 1 register clerk are present throughout the simulation period. As will be shown in section \ref{sec::Results}, the described input parameters of the base scenario led to typical and reasonable results in terms of waiting times, proportion of patient that meet \ac{TTD} targets and doctor utilization rates.

\subsection{Scenarios} \label{sec::Scenarios}

The data and parameters presented in section \ref{sec::data} defines the baseline scenario for experimentation. Hence, in the remainder it is referred to as ``Base''. However, to evaluate the performance of the proposed method under a variety of settings, additional scenarios based on alterations of some parameters are defined.

Parameters used for scenario definition include the arrival rate of patients, the variance of consultation times, and the distribution of patients over triage grades.

To test the \ac{ML} based selection method under the setting of a ``more severely'' crowded \ac{ED}, we included a scenario that is based on an 10\% increase of the hourly arrival rates shown by figure \ref{fig::ArrivalStream}. Note that the hourly pattern of arriving patients was left unchanged.

The variance of consultation times may influence the ``predictability'' of how long patients will occupy doctors during consultation. Consequently, the \ac{ML} based selection policy may suffer from high variances and vice-versa benefit from lower variances of consultation times. Hence, we define two additional scenarios with increased/decreased $\sigma$ parameters of the Log-Normal distributions for consultation times. Note that the corresponding $\mu$ parameters were adapted such that the expected duration of activities shown by table \ref{tab::ActivityDurations} remain unchanged.

Last, the distribution of patients over triage grades by \cite{Ferrand2018} assigns a relatively low probability mass to category 2 patients and almost a triangular shape to category 3 to 5 patients centered at category 4 patients. We define three alterations to this distribution. One, were category 3-5 patients have approximately uniformly distributed probabilities, and two where the peak of category 4 patients is shifted to either category 3 or 5 patients.

The resulting parameter sets and the scenario identifiers are summarized by table \ref{tab::ScenarioDefinition}.

For each scenario we randomly generated a set of 10,000 training instances and 10,000 evaluation instances. Training instances are used to twofold: (1) to compute near optimal solutions that are used to train ML models; (2) and to optimize weights $\beta$ for the \ac{APQ} algorithm. Each instance covers 24h hours starting at midnight. However, only patients arriving between 8am and 8pm (regardless of the time they leave the \ac{ED}) are used to compute performance measures and evaluate results. Hence, the time between midnight and 8am may be seen as a warm-up period of the models.

Besides different scenarios, we evaluate the proposed method with respect to a set of performance measures based on parameters $\lambda_W$, $\lambda_{ttd}$. The used objective functions include the ``pure'' ones, i.e. where only the waiting time or the \ac{TTD} target is considered, and two combined measures (indicated by the prefix C). For the combined measures one percent of target increase accounts for $\lambda_W$ minutes of waiting time reduction. All four combinations are summarized in Table \ref{tab::ScenarioDefinition}.

\begin{table}[]
\begin{center}

\begin{tabular}{lccc}
\hline
Scenario   Name & Arrival Rate Factor     & $\sigma$ & Patient Mix                  \\
\hline
Base           & 1                       & 0.45 & (0.065, 0.2, 0.515, 0.22)    \\
HR             & 1.1 & 0.45 & (0.065, 0.2, 0.515, 0.22)    \\
LV             & 1                       & 0.35 & (0.065, 0.2, 0.515, 0.22)    \\
HV             & 1                       & 0.55 & (0.065, 0.2, 0.515, 0.22)    \\
U              & 1                       & 0.45 & (0.065, 0.312, 0.312, 0.311) \\
T3             & 1                       & 0.45 & (0.065, 0.515, 0.2, 0.22)    \\
T5             & 1                       & 0.45 & (0.065, 0.2, 0.22, 0.515)    \\
\hline
Objective Name & $\lambda_{ttd}$                    & $\lambda_{W}$                        &      \\
\hline
TTDL           & 1    & 0    &                              \\
C-30           & 30   & 1    &                              \\
C-15           & 15   & 1    &                              \\
TWT            & 0    & 1    &                                             \\
\hline
\end{tabular}
\caption{Summary of Scenario and Objective Function Definitions}
\label{tab::ScenarioDefinition}
\end{center}
\end{table}

\section{Results} \label{sec::Results}

In this section we present the experimental results for all proposed scenario and objective function combinations. Algorithms and simulation models were coded in C\# and experiments were conducted on a desktop machine with an AMD Threadripper Pro 3995WX Processor containing 64 cores, each up to 4.2 GHz. Machine learning models were built using the Python scikit-learn 0.24.2 library \cite{scikit-learn} and then converted to C\# models. Each forest consists of 200 trees. To avoid over-fitting, each tree is based on a sample set consisting of 0.5 times the number of samples of the entire training set that are selected using bootstrapping. To account for imbalanced feature sets, e.g. in the base scenario the majority of patients are of category 4, we use over-sampling to maintain a balance of observed responses. However, we over-sample the first and second consultation responses individually. In particular, we choose the class for each type of consultation with the most responses, e.g. 41 for the first consultation in the base scenario, and for each other class we randomly bootstrap samples from the entire set belonging to this class until the same number of samples as for the maximum (in this example 41) is reached.

To parametrize and tune the Genetic Algorithm and the Simulated Annealing algorithm, we used the base scenario and the C-15 objective. Hence, the performance of those algorithms for other scenario/objective combinations, would possibly benefit from specific parametrization. In order to keep the computational effort at a reasonable level, we used the same versions of those algorithms for all settings. However, experiments have shown that the Simulated Annealing algorithms perform worse for the pure TTDL objective. This is mainly due to the approximately step-wise behavior of the objective function for a single day instance. In particular, the objective function only changes when a patient meets the target compared to a previous solution where the patient missed the target (and vice-versa). This results in many neighborhood moves not changing the objective value, which consequently limits the performance of the algorithm. Therefore, for computing optimal solutions we used a C-120 objective, i.e. $\lambda_{ttd} = 120$ and $\lambda_W = 1$, while presented results are evaluated by the TTDL objective.

The section is structured as follows. In Section \ref{sec::PredictiontResults} the prediction quality of the proposed ML models is evaluated. Section \ref{sec::BasetResults} outlines experimental results regarding the base scenario, while results for additional scenarios are summarized in Section \ref{sec::AdditionalResults}.

\subsection{Prediction Results}    \label{sec::PredictiontResults}

\acrodef{MCC}{Matthews Correlation Coefficient}

The patient selection method is based on a Random Forest classifier. To assess the quality of predictions compared to the optimal solutions, we further split the training instances (for which optimal solutions are known) into \ac{ML} training (90\%) and \ac{ML} test instances (10\% of all training instances). The fitted models are then evaluated on the \ac{ML} test instances. Overall results and results per triage category are summarized in Table \ref{tab::PredictionAccuracy}. For each included combination of scenarios and objective functions,  we report the accuracy of the predictor and the \ac{MCC}. We define accuracy as the number of correctly predicted observations of a class divided by the total number of predictions of that particular class. The \ac{MCC} is often used to assess classifiers when the underlying feature-response set is strongly unbalanced. It is bounded by $-1$ and $1$, where $1$ denotes a perfect prediction, $0$ a random prediction and $-1$ a prediction doing exactly the opposite to real world observations. Values above $0.4$ are considered to indicate a strong positive correlation between predictions and real world observations, and values above $0.7$ a very strong correlation.

\begin{table}[]
\begin{center}
\begin{tabular}{cc cc | cccc | cccc}
\hline
\multicolumn{1}{l}{}         &           &      &          &          & \multicolumn{7}{c}{Accuracy Per Triage   Grade}                       \\
\multicolumn{1}{l}{}         &           &      &          & \multicolumn{4}{c|}{First Consultation} & \multicolumn{4}{c}{Second Consultation} \\
\multicolumn{1}{l}{Scenario} & Objective & MCC  & Ov. Acc. & 2        & 3       & 4       & 5       & 2        & 3        & 4       & 5       \\
\hline
\multirow{4}{*}{Base} & TTDL & 0.71 & 77.3\% & 67.0\% & 67.6\% & 79.3\% & 77.6\% & 65.1\% & 82.0\% & 95.1\% & 86.6\% \\
                      & C-30 & 0.63 & 70.0\% & 65.9\% & 65.1\% & 78.6\% & 68.5\% & 45.7\% & 62.2\% & 74.0\% & 59.9\% \\
                      & C-15 & 0.63 & 70.0\% & 66.8\% & 65.6\% & 78.4\% & 67.3\% & 47.0\% & 61.4\% & 74.1\% & 60.4\% \\
                      & TWT  & 0.90 & 92.4\% & 89.6\% & 92.2\% & 96.6\% & 94.0\% & 85.6\% & 87.4\% & 90.8\% & 79.8\% \\
                      \hline
HR                    & \multirow{6}{*}{C-15}    & 0.56 & 64.6\% & 64.0\% & 64.3\% & 77.1\% & 56.2\% & 38.9\% & 52.7\% & 65.1\% & 48.1\% \\
LV                    &                          & 0.62 & 69.6\% & 68.0\% & 65.7\% & 78.3\% & 67.2\% & 46.4\% & 60.1\% & 73.1\% & 57.7\% \\
HV                    &                          & 0.62 & 69.9\% & 67.7\% & 65.4\% & 78.5\% & 67.8\% & 45.5\% & 61.4\% & 73.1\% & 59.0\% \\
U                     &                          & 0.62 & 68.6\% & 65.0\% & 71.2\% & 72.6\% & 75.7\% & 42.1\% & 64.9\% & 63.3\% & 62.1\% \\
T3                    &                          & 0.61 & 68.7\% & 63.6\% & 78.6\% & 66.1\% & 63.7\% & 44.9\% & 70.3\% & 52.1\% & 52.6\% \\
T5                    &                          & 0.70 & 75.8\% & 71.1\% & 71.5\% & 67.7\% & 86.5\% & 56.6\% & 71.4\% & 69.3\% & 83.4\%    \\
                             \hline
\end{tabular}
\caption{Prediction Accuracy and MCC for the \ac{ML} Classifier}
\label{tab::PredictionAccuracy}
\end{center}
\end{table}

Results show that for most of the settings the \ac{MCC} is above 0.6 with corresponding accuracy values larger than 68\% indicating a relatively strong correlation between predictions and optimal solutions. In particular, for the base scenario, results for the TDDL and TWT objectives (i.e. the single measures) are significantly higher, where the overall accuracy even exceeds 90\% for the TWT objective. Hence, it may be concluded that predicting optimal decisions is ``easier'' with respect to a single performance measure compared to a combined measure of at least partly conflicting objectives.

However, it has to be noted that a wrong prediction with respect to the optimal solution is not necessarily a bad decision for the given state of the \ac{ED}.

\subsection{Base Scenario Results} \label{sec::BasetResults}

In this section we analyze the results for the base scenario and different objectives. Table \ref{tab::ObjectivesBase} summarizes the estimated performance measure $\hat{f}_{ED}$ for all included combinations of $\lambda_W$ and $\lambda_{ttd}$ and patient selection policies. The selection policies include: the pure priority based policies QP1-4, the APQ method, and the \ac{ML} based method denoted by ML-OPT. Further, we report results based on the (near) optimal solutions (denoted by OPT) of the training instances as a benchmark. Thereby, it has to be noted that results for the OPT policy are based on different instances than the other policies. Further, it has to be noted that results for the APQ (and the ML-OPT) policy are based on different models for each objective function, but reported in a single line in table \ref{tab::ObjectivesBase}. In other words, the objective value for the C-15 measure and the ML-OPT policy results from the model that was fit on optimal solutions with respect to the C-15 objective. The same model would lead to a different result in terms of the C-30 objective than reported by Table \ref{tab::ObjectivesBase}, which itself is based on optimal solutions computed with respect to the C-30 objective.

\begin{table}[]
\begin{center}

\begin{tabular}{l cccc}
\hline
                     & \multicolumn{4}{c}{Objective Function} \\
                     \hline
Selection Policy     & TTDL     & C-30     & C-15    & TWT    \\
\hline
QP1                  & 7.65     & 445.3    & 330.6   & 215.9  \\
QP2                  & 17.22    & 649.6    & 391.3   & 132.9  \\
QP3                  & 31.43    & 1045.6   & 574.2   & 102.8  \\
QP4                  & 28.79    & 970.0    & 538.1   & 106.2  \\
\hline
Min QPX             & 7.65     & 445.3    & 330.6   & 102.8  \\
\hline
APQ                & 5.97     & 395.2    & 298.3   & 100.9  \\
ML-OPT             & 6.00     & 362.0    & 272.8   & 102.5  \\
OPT                & 4.89     & 316.0    & 243.4   & 102.4  \\
\hline
APQ-Improvement    & 21.9\%   & 11.3\%   & 9.8\%   & 1.8\%  \\
ML-OPT-Improvement & 21.6\%   & 18.7\%   & 17.5\%  & 0.3\%  \\
OPT-Improvement    & 36.1\%   & 29.0\%   & 26.4\%  & 0.4\%  \\

\hline
\end{tabular}
\caption{Summary of Objective Values $\hat{f}_{ED}$ and Improvement with Respect to Pure Priority-Based Queuing per Selection Policy for the Base Scenario}
\label{tab::ObjectivesBase}
\end{center}
\end{table}

Experiments show that for objective functions that are based on a single measure, i.e. either \ac{TTD} targets or TWT, \ac{APQ} and the ML-OPT approach lead to similar results. While the \ac{APQ} policy performs slightly better, both fail to improve the QP3 policy significantly for the TWT objective. For the TTDL objective, both methods yield an improvement above 21\%. In terms of pure priority based policies, the QP1 policy yields the best results for all objectives but for the TWT objective where QP3 performs best.

However, for objectives including both total waiting times and \ac{TTD} targets, the ML-OPT policy clearly outperforms the \ac{APQ} method. Hence, it may be concluded that the ML-OPT is more capable of balancing those (at least partly) conflicting objectives. To further analyze the behavior of the proposed policies, Tables \ref{tab::TTDLResultsBase} and \ref{tab::W1BaseResults} report expected \ac{TTD} target shares $E(X_{ttd}^j)$ and expected waiting time measures $E(X_{\tilde{W}_1})$ and $E(X_{\tilde{W}})$ per patient category $j$ and per policy. Further, the tables include 95\% confidence intervals of above mentioned measures to assess the accuracy of results.

\begin{table}[]
\begin{center}

\begin{tabular}{lccccc}
\hline
                & \multicolumn{5}{c}{Triage Category}                                                               \\
          
Selection Policy          & Overall           & 2                 & 3                 & 4                 & 5                 \\
\hline
QP1           & 83.7 $\pm$ 0.06\% & 99.0 $\pm$ 0.06\% & 75.9 $\pm$ 0.16\% & 90.3 $\pm$ 0.07\% & 70.9 $\pm$ 0.16\% \\
QP2           & 71.1 $\pm$ 0.08\% & 99.1 $\pm$ 0.06\% & 75.9 $\pm$ 0.16\% & 80.1 $\pm$ 0.09\% & 37.6 $\pm$ 0.17\% \\
QP3           & 61.4 $\pm$ 0.08\% & 96.1 $\pm$ 0.13\% & 71.9 $\pm$ 0.17\% & 65.0 $\pm$ 0.11\% & 33.2 $\pm$ 0.17\% \\
QP4           & 63.2 $\pm$ 0.08\% & 99.0 $\pm$ 0.06\% & 75.0 $\pm$ 0.16\% & 66.7 $\pm$ 0.11\% & 33.7 $\pm$ 0.17\% \\
\hline
APQ-TTDL    & 81.4 $\pm$ 0.07\% & 95.2 $\pm$ 0.14\% & 76.5 $\pm$ 0.16\% & 83.3 $\pm$ 0.09\% & 77.5 $\pm$ 0.15\% \\
ML-OPT-TTDL & 83.8 $\pm$ 0.10\% & 98.4 $\pm$ 0.16\% & 76.1 $\pm$ 0.14\% & 87.4 $\pm$ 0.04\% & 78.0 $\pm$ 0.42\% \\
OPT-TTDL    & 83.5 $\pm$ 0.08\% & 98.9 $\pm$ 0.27\% & 78.3 $\pm$ 0.14\% & 85.5 $\pm$ 0.03\% & 79.1 $\pm$ 0.33\% \\
\hline
APQ-C-30    & 81.2 $\pm$ 0.07\% & 94.5 $\pm$ 0.15\% & 76.2 $\pm$ 0.16\% & 83.4 $\pm$ 0.09\% & 76.8 $\pm$ 0.15\% \\
ML-OPT-C-30 & 83.6 $\pm$ 0.06\% & 98.2 $\pm$ 0.09\% & 76.1 $\pm$ 0.16\% & 87.3 $\pm$ 0.08\% & 77.5 $\pm$ 0.15\% \\
OPT-C-30    & 83.3 $\pm$ 0.06\% & 98.7 $\pm$ 0.07\% & 78.3 $\pm$ 0.15\% & 85.5 $\pm$ 0.08\% & 78.3 $\pm$ 0.15\% \\
\hline
APQ-C-15    & 81.2 $\pm$ 0.07\% & 95.1 $\pm$ 0.14\% & 76.3 $\pm$ 0.16\% & 83.4 $\pm$ 0.09\% & 76.5 $\pm$ 0.15\% \\
ML-OPT-C-15 & 83.5 $\pm$ 0.06\% & 98.2 $\pm$ 0.09\% & 76.2 $\pm$ 0.16\% & 87.4 $\pm$ 0.08\% & 76.5 $\pm$ 0.15\% \\
OPT-C-15    & 82.8 $\pm$ 0.06\% & 98.6 $\pm$ 0.08\% & 78.2 $\pm$ 0.16\% & 85.4 $\pm$ 0.08\% & 76.4 $\pm$ 0.15\% \\
\hline
APQ-TWT     & 51.6 $\pm$ 0.08\% & 95.8 $\pm$ 0.13\% & 51.7 $\pm$ 0.19\% & 52.6 $\pm$ 0.12\% & 36.2 $\pm$ 0.17\% \\
ML-OPT-TWT  & 61.4 $\pm$ 0.08\% & 96.0 $\pm$ 0.13\% & 71.8 $\pm$ 0.17\% & 65.0 $\pm$ 0.11\% & 33.3 $\pm$ 0.17\% \\
OPT-TWT     & 61.7 $\pm$ 0.08\% & 91.3 $\pm$ 0.19\% & 67.9 $\pm$ 0.18\% & 66.0 $\pm$ 0.11\% & 37.3 $\pm$ 0.17\% \\
\hline
\end{tabular}
\caption{Estimated Share of Patients Meeting \ac{TTD} target and 95\% Confidence Intervals per Selection Policy and Triage Grade for the Base Scenario}
\label{tab::TTDLResultsBase}
\end{center}
\end{table}

Table \ref{tab::TTDLResultsBase} shows that the performance of the \ac{APQ} method and the ML-OPT policy are relatively similar in terms of \ac{TTD} target shares for all objectives, except for the TWT, which is based only on waiting times. Interestingly, both the OPT and ML-OPT policy yield a target rate above 98\% for patients of category 2 which is about 3\% higher than the target rate of 95\%. In other words, those 3\% do not contribute to the improvement of those methods compared to the \ac{APQ} policy. Similarly for category 4 patients, the ML-OPT policy exceeds the target share of 83\%. For the TWT objective, the performance of the ML-OPT policy is similar to that of the best pure priority based policy, QP3, in terms of \ac{TTD} target shares, while applying the \ac{APQ} policy results in significantly lower shares of patients meeting target times for all but category 5 patients.

\begin{table}[]
\begin{center}
\begin{tabular}{lccccc}
\hline
              & \multicolumn{5}{c}{Triage Category}                                                     \\
Selection Policy        & Overall         & 2               & 3               & 4               & 5               \\
\hline
              & \multicolumn{5}{c}{$E(X_{\tilde{W}_1})$}                                                                   \\
\hline
QP1         & 33.7 $\pm$ 0.1  & 3.1 $\pm$ 0.0   & 13.7 $\pm$ 0.1  & 19.9 $\pm$ 0.1  & 80.6 $\pm$ 0.4  \\
QP2         & 73.6 $\pm$ 0.3  & 3.0 $\pm$ 0.0   & 13.7 $\pm$ 0.1  & 28.9 $\pm$ 0.1  & 217.4 $\pm$ 0.8 \\
QP3         & 89.2 $\pm$ 0.3  & 4.1 $\pm$ 0.0   & 15.4 $\pm$ 0.1  & 45.5 $\pm$ 0.1  & 241.9 $\pm$ 0.8 \\
QP4         & 87.4 $\pm$ 0.3  & 3.0 $\pm$ 0.0   & 14.1 $\pm$ 0.1  & 43.7 $\pm$ 0.1  & 239.8 $\pm$ 0.8 \\
\hline
APQ-TTDL    & 32.7 $\pm$ 0.1  & 5.0 $\pm$ 0.0   & 13.5 $\pm$ 0.1  & 25.9 $\pm$ 0.1  & 63.5 $\pm$ 0.3  \\
ML-OPT-TTDL & 38.7 $\pm$ 0.1  & 4.5 $\pm$ 0.1   & 16.9 $\pm$ 0.2  & 27.8 $\pm$ 0.1  & 80.7 $\pm$ 0.2  \\
OPT-TTDL    & 40.9 $\pm$ 0.0  & 5.6 $\pm$ 0.0   & 21.9 $\pm$ 0.2  & 30.5 $\pm$ 0.1  & 80.8 $\pm$ -0.2 \\
\hline
APQ-C-30    & 34.8 $\pm$ 0.1  & 5.4 $\pm$ 0.1   & 13.7 $\pm$ 0.1  & 26.9 $\pm$ 0.1  & 69.7 $\pm$ 0.3  \\
ML-OPT-C-30 & 39.0 $\pm$ 0.1  & 4.8 $\pm$ 0.2   & 16.8 $\pm$ 0.1  & 28.0 $\pm$ 0.1  & 81.6 $\pm$ 0.3  \\
OPT-C-30    & 41.5 $\pm$ 0.1  & 5.8 $\pm$ 0.1   & 21.9 $\pm$ 0.2  & 30.6 $\pm$ 0.1  & 83.0 $\pm$ 0.4  \\
\hline
APQ-C-15    & 35.1 $\pm$ 0.1  & 5.2 $\pm$ 0.0   & 13.7 $\pm$ 0.1  & 27.0 $\pm$ 0.1  & 70.5 $\pm$ 0.3  \\
ML-OPT-C-15 & 39.2 $\pm$ 0.1  & 4.7 $\pm$ 0.2   & 16.4 $\pm$ 0.1  & 28.0 $\pm$ 0.1  & 82.9 $\pm$ 0.4  \\
OPT-C-15    & 43.0 $\pm$ 0.1  & 5.7 $\pm$ 0.1   & 21.9 $\pm$ 0.2  & 30.9 $\pm$ 0.1  & 88.1 $\pm$ 0.4  \\
\hline
APQ-TWT     & 81.8 $\pm$ 0.2  & 4.7 $\pm$ 0.0   & 22.0 $\pm$ 0.1  & 54.0 $\pm$ 0.1  & 190.7 $\pm$ 0.7 \\
ML-OPT-TWT  & 88.6 $\pm$ 0.3  & 4.6 $\pm$ 0.1   & 15.5 $\pm$ 0.1  & 45.4 $\pm$ 0.1  & 239.6 $\pm$ 0.8 \\
OPT-TWT     & 83.2 $\pm$ 0.3  & 7.6 $\pm$ 0.3   & 21.4 $\pm$ 0.2  & 43.9 $\pm$ 0.1  & 218.1 $\pm$ 0.8 \\
\hline
              & \multicolumn{5}{c}{$E(X_{\tilde{W}})$}                                                                   \\
\hline
QP1         & 295.1 $\pm$ 0.6 & 198.4 $\pm$ 1.5  & 241.6 $\pm$ 1.0 & 316.6 $\pm$ 0.8 & 376.0 $\pm$ 1.6 \\
QP2         & 154.9 $\pm$ 0.5 & 14.1 $\pm$ 0.1   & 31.1 $\pm$ 0.1  & 167.5 $\pm$ 0.6 & 395.6 $\pm$ 1.6 \\
QP3         & 77.5 $\pm$ 0.3  & 14.3 $\pm$ 0.1   & 26.1 $\pm$ 0.1  & 57.0 $\pm$ 0.2  & 252.1 $\pm$ 1.3 \\
QP4         & 88.8 $\pm$ 0.3  & 13.1 $\pm$ 0.1   & 25.7 $\pm$ 0.1  & 62.6 $\pm$ 0.2  & 304.5 $\pm$ 1.2 \\
\hline
APQ-TTDL    & 290.5 $\pm$ 0.5 & 190.5 $\pm$ 1.3  & 243.4 $\pm$ 0.9 & 311.2 $\pm$ 0.8 & 365.6 $\pm$ 1.5 \\
ML-OPT-TTDL & 227.8 $\pm$ 0.2 & 168.9 $\pm$ -3.9 & 204.4 $\pm$ 3.2 & 235.2 $\pm$ 0.4 & 278.7 $\pm$ 1.0 \\
OPT-TTDL    & 200.1 $\pm$ 1.2 & 185.9 $\pm$ -2.0 & 189.7 $\pm$ 2.2 & 192.0 $\pm$ 2.1 & 247.5 $\pm$ 1.4 \\
\hline
APQ-C-30    & 268.8 $\pm$ 0.5 & 241.1 $\pm$ 1.3  & 254.2 $\pm$ 0.9 & 256.5 $\pm$ 0.6 & 342.1 $\pm$ 1.4 \\
ML-OPT-C-30 & 225.9 $\pm$ 0.4 & 180.1 $\pm$ 1.4  & 205.1 $\pm$ 0.9 & 229.2 $\pm$ 0.6 & 276.2 $\pm$ 1.2 \\
OPT-C-30    & 197.8 $\pm$ 0.4 & 186.8 $\pm$ 1.5  & 185.9 $\pm$ 0.9 & 188.9 $\pm$ 0.6 & 247.5 $\pm$ 1.2 \\
\hline
APQ-C-15    & 269.9 $\pm$ 0.5 & 218.4 $\pm$ 1.3  & 246.2 $\pm$ 0.8 & 260.6 $\pm$ 0.7 & 363.4 $\pm$ 1.5 \\
ML-OPT-C-15 & 225.9 $\pm$ 0.4 & 175.4 $\pm$ 1.4  & 202.4 $\pm$ 0.8 & 229.8 $\pm$ 0.6 & 281.5 $\pm$ 1.2 \\
OPT-C-15    & 192.0 $\pm$ 0.4 & 178.2 $\pm$ 1.5  & 177.7 $\pm$ 0.8 & 182.5 $\pm$ 0.6 & 248.9 $\pm$ 1.2 \\
\hline
APQ-TWT     & 74.7 $\pm$ 0.2  & 14.9 $\pm$ 0.1   & 32.6 $\pm$ 0.1  & 65.7 $\pm$ 0.2  & 201.1 $\pm$ 1.0 \\
ML-OPT-TWT  & 77.4 $\pm$ 0.3  & 14.7 $\pm$ 0.1   & 26.3 $\pm$ 0.1  & 57.0 $\pm$ 0.2  & 250.7 $\pm$ 1.2 \\
OPT-TWT     & 82.2 $\pm$ 0.3  & 25.9 $\pm$ 0.5   & 38.2 $\pm$ 0.3  & 59.9 $\pm$ 0.2  & 246.2 $\pm$ 1.2 \\
\hline
\end{tabular}
\caption{Estimated Total Waiting Times $E(X_{\tilde{W}_1}$ and $E(X_{\tilde{W}}$  and 95\% Confidence Intervals per Selection Policy and Triage Grade for the Base Scenario}
\label{tab::W1BaseResults}
\end{center}
\end{table}

Comparing expected total waiting times for patients leaving the \ac{ED} after the first consultation ($E(X_{\tilde{W}_1})$) and patients requiring a second consultation ($E(X_{\tilde{W}})$) reveals the major difference in behavior of the \ac{APQ} and the ML-OPT policies. For all objectives that include \ac{TTD} target shares (i.e. TTDL, C-15 and C-30), ML-OPT (and also OPT) results in longer expected waiting times than \ac{APQ} for patients of categories 3-5 that leave the \ac{ED} after the first consultation. Especially for patients of category 5 the difference is significant. However, ML-OPT (and also) OPT significantly reduce the expected total waiting time $E(X_{\tilde{W}})$ for patients requiring a second consultation regardless of the triage category.

In particular, the results of the \ac{APQ} policy with respect to objectives C-15 and C-30 reveal only a slight increase in terms of $E(X_{\tilde{W}})$ over patient categories 2-4, but significantly longer waiting times for patients of category 5. This is especially interesting, as the \ac{APQ} method is designed to balance waiting times between high and low priority patients, but fails to do so for patients of very low priority, i.e. patients of category 5. On the other hand, for C-15 and C-30 objectives, the ML-OPT method results in a steeper curve of $E(X_{\tilde{W}})$ values over patient categories but on a generally lower level than the \ac{APQ} policy . Hence, one may conclude that ML-OPT is better suited to balance conflicting \ac{TTD} target shares and expected total waiting times for all patient categories. This may result from the observation that \ac{APQ} weight vectors $\beta$ for C-15 and C-30 only assign about 1.1\% of their mass to second consultation patients (see table \ref{tab::APQWeights}), i.e. \ac{APQ} is only rarely mixing ``new'' and ``old''. Clearly, the ML-OPT method results in a more balanced mix of patients waiting for the first and second consultation.

For the pure target-based objective TTDL, the results for \ac{APQ} in terms of $E(X_{\tilde{W}_1})$ are similar to the results for the combined measures. This can be explained by the observations that $E(X_{\tilde{W}})$ values more or less correlate with \ac{TTD} target shares for those patients. However, results regarding $E(X_{\tilde{W}})$ show significantly longer waiting times for patients with low priority that require a second consultation. This could have been expected as those waiting times do not contribute to the TTDL objective, and consequently the sum of weights in vector $\beta$ for second consultation patients is about 0.3\% (see table \ref{tab::APQWeights}). The performance of the ML-OPT method is relatively similar  for the TTDL and C-30 objectives. This may be because the computation of the optimal solutions was done using a C-120 objective rather than the pure TTDL objective, when training the ML models.

Results for the pure TWT objective show that both policies, i.e. \ac{APQ} and ML-OPT, result in waiting times for category 5 patients significantly longer than for patients of higher priority. However, it has to be noted that \ac{APQ} does slightly better than QP3, while ML-OPT performs on a similar level compared to the best pure priority based policy.

\subsection{Additional Scenarios Results} \label{sec::AdditionalResults}

In this section, we present results for scenarios with altered parameters. All results were obtained by applying the C-15 objective, i.e. a mix of \ac{TTD} target shares and expected total waiting times.

Table \ref{tab::ObjectivesAdditional} shows the objective values and improvement with respect to pure priority based policies for all additional scenarios.

\begin{table}[]
\begin{center}
    
\begin{tabular}{l cccccc}
\hline
                     & \multicolumn{6}{c}{Scenario}                                                                                                                       \\

Selection Policy     & \multicolumn{1}{c}{HR} & \multicolumn{1}{c}{LV} & \multicolumn{1}{c}{HV} & \multicolumn{1}{c}{U} & \multicolumn{1}{c}{T3} & \multicolumn{1}{c}{T5} \\
\hline
QP1                  & 669.3                  & 324.6                  & 340.7                  & 327.5                 & 706.6                  & 232.9                  \\
QP2                  & 795.2                  & 375.5                  & 416.3                  & 375.8                 & 812.1                  & 256.4                  \\
QP3                  & 977.8                  & 564.6                  & 600.0                  & 519.0                 & 1056.2                 & 333.6                  \\
QP4                  & 946.5                  & 526.3                  & 558.0                  & 474.8                 & 999.7                  & 317.8                  \\
\hline
Min QPX             & 669.3                  & 324.6                  & 340.7                  & 327.5                 & 706.6                  & 232.9                  \\
\hline
APQ                & 635.5                  & 291.1                  & 319.4                  & 369.0                 & 704.7                  & 180.4                  \\
ML-OPT             & 584.4                  & 263.0                  & 286.5                  & 342.8                 & 659.3                  & 193.7                  \\
OPT                & 516.3                  & 244.3                  & 250.2                  & 284.6                 & 546.6                  & 164.7                  \\
\hline
APQ-Improvement    & 5.1\%                  & 10.3\%                 & 6.2\%                  & 5.5\%                 & 0.3\%                  & 22.5\%                 \\
ML-OPT-Improvement & 12.7\%                 & 19.0\%                 & 15.9\%                 & 12.2\%                & 6.7\%                  & 16.8\%                 \\
OPT-Improvement    & 22.9\%                 & 24.7\%                 & 26.5\%                 & 27.1\%                & 22.6\%                 & 29.3\%     \\
\hline        
\end{tabular}
\caption{Summary of Objective Values $\hat{f}_{ED}$ and Improvement with Respect to Pure Priority Based Queuing  per Selection Policy for the Additional Scenarios}
\label{tab::ObjectivesAdditional}
\end{center}
\end{table}

Results for the HR scenario, i.e. an increased number of arriving patients, show that the improvement of both the \ac{APQ} and the ML-OPT policy is less than for the base scenario. In other words, for a highly crowded \ac{ED} there is less potential for improvement than in the more typical setting of the base scenario. However, it has to be noted that the crowded setting impacts the improvement of the \ac{APQ} method to a larger extent (i.e. from 9.8\% to 5.1\%) than it impacts the ML-OPT method (i.e. from 17.5\% to 12.7\%). Comparing the values of the (not realistic) OPT policy shows that there generally a lower potential improvement that theoretically could be achieved, as the improvement drops from 26.4\% to 22.9\%.

Under scenarios of higher and lower variance in the  consultation time distributions, i.e. LV and HV, the methods perform as expected. A lower variance in consultation times leads to an increased improvement for both the \ac{APQ} and the ML-OPT method, and the improvement decreases for a setting with higher consultation time variances. However, it is interesting to observe that the \ac{APQ} method seems to be impacted more than the ML-OPT method by increasing stochasticity in the system.

The performance of the methods is more differentiated for scenarios that are based on altered patient mixes. ML-OPT clearly outperforms \ac{APQ} for scenario U, where the probability of a patient belonging to triage categories 3-5 is almost constant. However, the gap between ML-OPT and the theoretical optimum OPT is significantly larger than for the base scenario. A possible explanation could be, that for uniformly distributed patient arrivals of category 3-5, at the time of decision making the variability of future states of the \ac{ED} is larger than for patient mixes with a dominant patient category.

The T3 scenario, i.e. a majority of patients arriving belong to category 3, leads to significantly less improvement of all policies compared to QP1. In particular, \ac{APQ} does not yield any significant improvement, while ML-OPT only improves the overall objective by 6.7\%. Besides a low number of category 2 patients, the majority of patients have very high priority in this scenario. This mix is even more extreme for second consultation queues, as category 3 patients more often require diagnostic tests and consequently a second consultation. Hence, queues of category 3 patients will be most ``crowded'', which leaves less potential for improvement.

The \ac{APQ} policy outperforms ML-OPT for scenario T5, i.e., the majority of patients belonging to category 5. In some sense this scenario is the extreme case of what motivates the APQ approach, with most patients having the lowest priority, and so it is not surprising this method performs well.

To further analyse the behavior of the proposed policies for the additional scenarios, Tables \ref{tab::TTDLResultsAdditional} and \ref{tab::W1AdditionalResults} report expected \ac{TTD} target shares $E(X_{ttd}^j)$ and expected waiting time measures $E(X_{\tilde{W}_1})$ and $E(X_{\tilde{W}})$ per patient category $j$.

\begin{table}[]
\begin{center}
    
\begin{tabular}{cl ccccc}
\hline
\multicolumn{1}{l}{} &          & \multicolumn{5}{c}{Triage Category} \\
Scenario & Selection Policy         & Overall              & 2       & 3       & 4      & 5      \\
\hline
\multirow{4}{*}{HR}  & Min QPX & 69.5\%               & 98.9\%  & 59.6\%  & 77.9\% & 50.0\% \\
                     & APQ      & 64.1\%               & 95.3\%  & 60.3\%  & 78.5\% & 24.3\% \\
                     & ML-OPT & 70.3\%               & 97.5\%  & 59.6\%  & 77.6\% & 54.9\% \\
                     & OPT      & 72.4\%               & 97.3\%  & 63.1\%  & 78.9\% & 58.2\% \\
\hline
\multirow{4}{*}{LV}  & Min QPX & 84.5\%               & 97.3\%  & 76.4\%  & 91.6\% & 71.5\% \\
                     & APQ      & 81.8\%               & 93.0\%  & 77.0\%  & 83.7\% & 78.2\% \\
                     & ML-OPT & 84.3\%               & 96.6\%  & 76.7\%  & 88.4\% & 77.9\% \\
                     & OPT      & 82.9\%               & 98.8\%  & 78.4\%  & 85.3\% & 76.5\% \\
\hline
\multirow{4}{*}{HV}  & Min QPX & 82.3\%               & 96.7\%  & 75.2\%  & 88.6\% & 69.9\% \\
                     & APQ      & 80.3\%               & 93.9\%  & 75.5\%  & 83.2\% & 73.7\% \\
                     & ML-OPT & 81.7\%               & 95.2\%  & 75.3\%  & 85.7\% & 74.4\% \\
                     & OPT      & 82.4\%               & 98.0\%  & 77.7\%  & 85.3\% & 75.0\% \\
\hline
\multirow{4}{*}{U}   & Min QPX & 81.9\%               & 99.0\%  & 74.5\%  & 91.4\% & 76.2\% \\
                     & APQ      & 79.7\%               & 93.2\%  & 74.7\%  & 82.5\% & 79.2\% \\
                     & ML-OPT & 80.2\%               & 98.7\%  & 74.7\%  & 85.7\% & 76.5\% \\
                     & OPT      & 82.0\%               & 98.8\%  & 78.3\%  & 85.9\% & 78.1\% \\
\hline
\multirow{4}{*}{T3}  & Min QPX & 71.9\%               & 98.6\%  & 70.7\%  & 77.9\% & 61.5\% \\
                     & APQ      & 71.3\%               & 93.3\%  & 70.9\%  & 77.5\% & 59.9\% \\
                     & ML-OPT & 71.4\%               & 97.3\%  & 70.1\%  & 76.5\% & 62.3\% \\
                     & OPT      & 74.1\%               & 95.7\%  & 75.6\%  & 75.6\% & 62.8\% \\
\hline
\multirow{4}{*}{T5}  & Min QPX & 90.8\%               & 99.3\%  & 76.2\%  & 96.1\% & 93.2\% \\
                     & APQ      & 81.2\%               & 94.9\%  & 77.0\%  & 83.5\% & 80.2\% \\
                     & ML-OPT & 86.6\%               & 99.1\%  & 76.5\%  & 94.5\% & 85.5\% \\
                     & OPT      & 88.5\%               & 99.6\%  & 79.9\%  & 90.9\% & 89.5\% \\         
\hline
\end{tabular}
\caption{Estimated Share of Patients Meeting \ac{TTD} target per Selection Policy and Triage Grade for Additional Scenarios}
\label{tab::TTDLResultsAdditional}
\end{center}
\end{table}

As could have been expected, an increased arrival rate of patients decreases the number of patients meeting \ac{TTD} target times. While results for category 2 patients remain at a similar level compared to the base scenario, patients of lower priority meet their target times less often. However, a large difference between \ac{APQ} and ML-OPT can only be observed for patients of category 5, where ML-OPT significantly outperforms \ac{APQ}.

In a similar fashion, altered consultation time variances seem to impact category 5 patients the most, when comparing \ac{APQ} and ML-OPT results.

For scenarios based on adapted patient mixes, only marginal differences below corresponding thresholds can be observed for patients with priority 2 or 3 (see Table \ref{tab::TTDLResultsAdditional}). For scenario U, ML-OPT prioritises category 4 patients, leading to worse performance for category 5 patients compared to \ac{APQ}. For scenario T3 the ML-OPT policy leads to better results for category 5 patients, while only performing slightly worse for category 4 patients. Interestingly, ML-OPT  prioritises category 4 patients in terms of TTD target shares for the T5 scenario, but also increases the share of category 5 patients meeting target times. However, it achieves this increase by ``trading off against'' total waiting times, as shown in Table \ref{tab::W1AdditionalResults}.

\begin{table}[]
\begin{center}
    
\begin{tabular}{c l | ccccc | ccccc}
\hline
\multicolumn{1}{l}{} &          & \multicolumn{5}{c|}{$E(X_{\tilde{W}_1})$}              & \multicolumn{5}{c}{$E(X_{\tilde{W}})$}                   \\
\hline
\multicolumn{1}{l}{} &          & \multicolumn{5}{c|}{Triage Category} & \multicolumn{5}{c}{Triage Category}     \\
\multicolumn{1}{l}{} &          & Overall & 2   & 3    & 4    & 5     & Overall & 2     & 3     & 4     & 5     \\
                     
                     \hline
\multirow{4}{*}{HR} & Min QPX & 57.4  & 3.4 & 21.6 & 30.9 & 145.0 & 437.4 & 323.2 & 377.1 & 462.9 & 527.2 \\
                    & APQ     & 110.4 & 5.1 & 21.2 & 31.3 & 351.4 & 232.4 & 172.8 & 194.0 & 211.1 & 386.9 \\
                    & ML-OPT  & 61.9  & 6.8 & 30.0 & 37.3 & 143.2 & 312.5 & 275.1 & 295.7 & 311.6 & 363.2 \\
                    & OPT     & 67.8  & 7.3 & 39.0 & 41.6 & 151.2 & 275.8 & 252.3 & 255.6 & 263.3 & 356.2 \\
\hline
\multirow{4}{*}{LV} & Min QPX & 32.6  & 3.1 & 13.5 & 18.9 & 78.4  & 295.0 & 198.3 & 241.0 & 316.8 & 376.2 \\
                    & APQ     & 34.6  & 5.2 & 13.4 & 27.3 & 68.1  & 265.5 & 201.4 & 252.4 & 254.7 & 354.2 \\
                    & ML-OPT  & 38.8  & 5.4 & 16.1 & 27.7 & 81.7  & 223.0 & 174.0 & 201.1 & 226.0 & 278.0 \\
                    & OPT     & 43.2  & 5.3 & 22.2 & 31.1 & 88.1  & 195.2 & 182.2 & 180.9 & 185.2 & 253.0 \\
\hline
\multirow{4}{*}{HV} & Min QPX & 35.4  & 3.2 & 13.9 & 21.2 & 84.3  & 296.9 & 202.0 & 245.4 & 317.3 & 376.8 \\
                    & APQ     & 36.5  & 4.6 & 13.8 & 26.8 & 76.3  & 273.6 & 221.3 & 245.0 & 277.5 & 338.7 \\
                    & ML-OPT  & 41.0  & 7.5 & 17.0 & 28.7 & 87.4  & 225.2 & 175.1 & 206.1 & 228.2 & 276.4 \\
                    & OPT     & 43.9  & 5.7 & 21.9 & 30.9 & 91.4  & 190.4 & 172.5 & 174.3 & 182.9 & 247.3 \\
\hline
\multirow{4}{*}{U}  & Min QPX & 34.7  & 3.1 & 14.2 & 18.9 & 66.9  & 301.0 & 197.8 & 258.1 & 333.3 & 378.7 \\
                    & APQ     & 37.2  & 5.6 & 14.3 & 27.8 & 65.0  & 271.0 & 162.9 & 255.7 & 303.4 & 303.9 \\
                    & ML-OPT  & 43.2  & 3.9 & 15.4 & 30.4 & 78.7  & 222.6 & 161.1 & 197.1 & 237.0 & 275.1 \\
                    & OPT     & 45.3  & 5.4 & 18.6 & 30.2 & 82.1  & 194.0 & 174.1 & 168.4 & 197.4 & 242.5 \\
\hline
\multirow{4}{*}{T3} & Min QPX & 45.0  & 3.4 & 15.9 & 33.5 & 107.6 & 364.7 & 243.0 & 336.5 & 439.8 & 468.8 \\
                    & APQ     & 47.0  & 5.6 & 15.8 & 34.6 & 113.7 & 340.3 & 239.1 & 344.8 & 351.5 & 370.6 \\
                    & ML-OPT  & 50.7  & 5.9 & 17.6 & 44.9 & 115.4 & 277.7 & 253.7 & 256.3 & 315.2 & 336.5 \\
                    & OPT     & 59.2  & 7.6 & 18.8 & 59.7 & 131.6 & 237.2 & 214.5 & 208.7 & 280.0 & 319.2 \\
\hline
\multirow{4}{*}{T5} & Min QPX & 25.8  & 2.8 & 13.4 & 14.5 & 35.1  & 232.4 & 153.9 & 191.1 & 234.1 & 278.9 \\
                    & APQ     & 44.8  & 5.2 & 13.2 & 27.5 & 62.6  & 146.2 & 17.4  & 66.9  & 150.4 & 229.4 \\
                    & ML-OPT  & 39.3  & 3.4 & 13.9 & 19.5 & 56.2  & 162.5 & 74.5  & 109.9 & 130.7 & 235.8 \\
                    & OPT     & 39.0  & 4.2 & 16.9 & 23.3 & 53.4  & 133.6 & 101.6 & 109.1 & 128.1 & 160.8  \\
                     \hline
\end{tabular}
\caption{Estimated Total Waiting Times $E(X_{\tilde{W}_1}$ and $E(X_{\tilde{W}}$ per Selection Policy and Triage Grade for the Additional Scenarios}
\label{tab::W1AdditionalResults}
\end{center}
\end{table}

Table \ref{tab::W1AdditionalResults} summarizes expected total waiting times for all additional scenarios and the C-15 objective. Investigating results for the HR scenario reveals some interesting behavior. While for the base scenario, the improvement of the ML-OPT method compared to \ac{APQ} mainly resulted from a reduction in $E(X_{\tilde{W}})$, results for HR differ. $E(X_{\tilde{W}})$ for this scenario is higher for patient categories 2-4 with the ML-OPT method than with the \ac{APQ} method, but ML-OPT performs significantly better for category 5 patients. Consequently, it may be concluded that the improvement of ML-OPT mainly results from ``not neglecting'' those low priority patients.

Observed waiting time behavior for scenarios LV and HV is fairly similar compared to the base scenario and objective C-15. For scenarios U and T3 the improvement of ML-OPT seems to be due to the decreased $E(X_{\tilde{W}})$ compared to the \ac{APQ} policy. This may lead to the same conclusion as for the base scenario, i.e. that ML-OPT is able to achieve a improved mix of ``new'' and ``old'' patients.

As outlined previously, the \ac{APQ} method outperforms the ML-OPT method for the T5 scenario. Analysing $E(X_{\tilde{W}_1})$ and $E(X_{\tilde{W}})$ measures for OPT reveals that the theoretical optimal policy balances $E(X_{\tilde{W}})$ over all triage grades, but at a significantly higher level than \ac{APQ}. While this leads to a theoretical improvement with respect to the C-15 objective, ML-OPT fails to imitate this behavior, especially for category 5 patients. Hence, it seems that the general ML model that was parametrized using the base scenario, is not capable of sufficiently capturing the dynamics of this particular scenario.

To sum up, the ML-OPT approach outperforms the \ac{APQ} policy for a majority of settings where the performance metric is based on \ac{TTD} targets and total expected waiting times. In most cases, the performance gain is yielded by an improved mix of first and second consultation patients.

\section{Conclusion and Further Research} \label{sec::Conclusion} 

We proposed a learning-based patient selection method for assigning idle resources, i.e. doctors, to waiting patients in an \ac{ED} setting. Once a doctor becomes idle, the policy picks a patient at the head of eight possible queues, one for each triage category and type of consultation. Thereby, it aims to imitate the behavior of (near) optimal solutions computed by a Simulated Annealing algorithm. In particular, the Simulated Annealing algorithm is used to heuristically optimize a large set of training instances, each covering a single typical day of the \ac{ED} setting. Thereby, the instances are treated as a deterministic input, with a priori known patient arrivals, categories, diagnostic requirements and consultation times. From this ``data-base'' of optimal solutions, a feature set is created, which should comprehend information on optimal decisions with respect to specific states of the \ac{ED}. This feature set is used to fit a standard machine learning classifier that builds the basis for the proposed selection policy. The method is compared to established patient selection policies, such as the \ac{APQ} method. Mathematically speaking, the main difference between the proposed method and \ac{APQ} is that our approach is based on a more comprehensive state representation as a basis for decision making, and a more dynamic and non-linear selection function, i.e. the \ac{ML} model. The proposed method is evaluated using a Discrete Event Simulation Model which imitates typical \ac{ED} settings that can be found in literature. Experiments show that the ML based policy outperforms the \ac{APQ} method in a majority of settings where the performance of the \ac{ED} is assessed by a combination of total waiting times and the share of patients meeting \ac{TTD} target times.

Results indicate that the proposed \ac{ML} based selection method is able to generate an improved mix of selected patients waiting to be seen either the first or the second time by a doctor. This results in lower overall waiting times, while keeping the share of patients meeting \ac{TTD} targets at a comparable level.

The presented method is evaluated using a typical \ac{ED} setting that is not based on a particular real world \ac{ED}. Hence, future research should include the transfer of the theoretical approach to a real world setting. Obviously, any such setting must include a computerized \ac{ED} management system, that could incorporate such a \ac{ML} based policy. Further, a simulation model of the real system is required for training of the \ac{ML} model. Although, this imposes some barriers for the practical use of the method, the same holds for the \ac{APQ} method, as one may doubt that priority scores will ever be computed manually to select patients. In addition, unless choosing weights arbitrarily, a simulation model based weight optimization is also required for the \ac{APQ} method.

Further, the presented study aims to demonstrate the theoretical potential of rather complex patient selection policies based on ML models. When transferring the approach to a real world setting, it would be interesting to investigate, if such complex policies could be used to derive simpler rule sets that are more easily interpretable, but extend the strictly linear logic of \ac{APQ}. In other words, future research should include the design and evaluation of models that cover the ``middle ground'' between simple \ac{APQ} policies and ``black box'' methods such as \ac{ML} models. However, the results presented in this paper, based on a ``black box'' method, clearly demonstrate that there is a significant improvement potential for the management of \ac{ED} resources that can be achieved using more sophisticated state representations and selection functions.


\bibliographystyle{abbrv}

\bibliography{literature}

\newpage

\appendix
\section{Appendix}
 
\subsection{\ac{APQ} Weight Optimization} \label{sec::APQOpt}

The implementation of \ac{APQ} patient selection requires the definition of weights $ \beta = (\beta_{j,k})_{j \in J, 1 \leq k \leq 2}$. \cite{Cildoz2019} use the OptQuest simulation-optimization package within the Arena simulation tool \cite{Laguna1997} to optimize weights $\beta$. In this paper, we use a genetic algorithm to optimize weights $\beta$ for a training set of simulation instances. Note that the training set may differ over scenarios, which have been defined in Section \ref{sec::Scenarios}, but the main optimization procedures remain the same and will be outlined in the following. For a general description of genetic algorithms the reader is referred to \cite{Whitley1994}.

\begin{figure}[ht!]
    \begin{center}
    \includegraphics[width=0.8\textwidth]{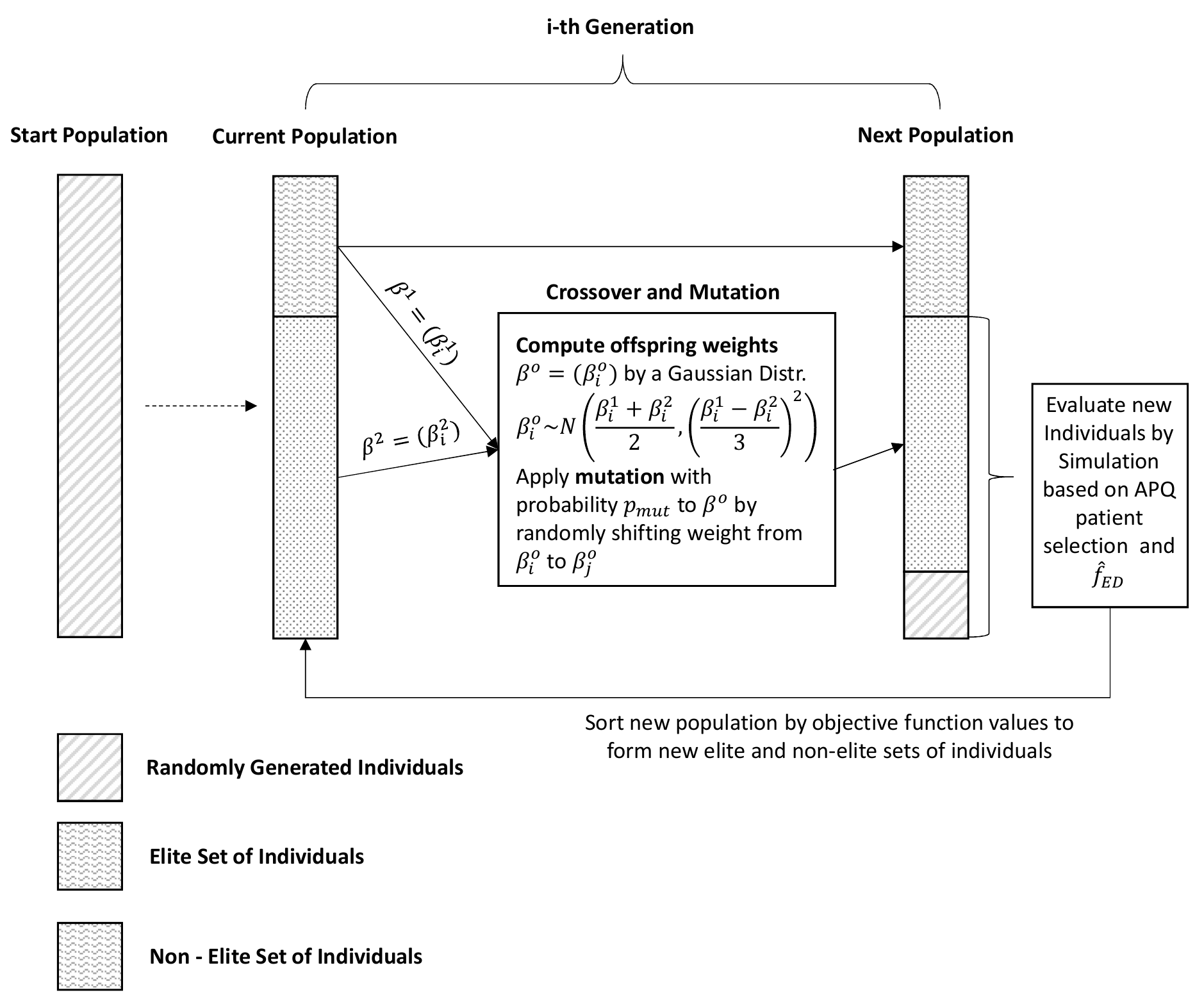}
    \caption{General Principle of the Genetic Algorithm Used to Optimize \ac{APQ} Weights $\beta$}
    \label{fig:GeneticAlgorithm}
    \end{center}
    
\end{figure}

The main principle of the algorithm is outlined in Figure \ref{fig:GeneticAlgorithm}. 

The algorithm is initialized with a set of randomly generated start solutions, i.e. randomly generated weight vectors $\beta$. 
To evaluate the fitness of an individual, a set of patients and associated data (i.e. arrival times, triage grades, activity durations and diagnostic requirements) is fed to the simulated model described in section \ref{sec::EDModel}. The \ac{APQ} method is used as a patient selection mechanism and the objective value $\hat{f}_{ED}$ is computed based on the data stored by the simulation.

The population at each generation during the execution of the genetic algorithm is divided into an elite-set of individuals (those with lowest objective value) and a non-elite-set of individuals (the remaining individuals). The newly formed population consists of the current elite-set, a set of off-spring generated by a crossover operator and randomly generated mutant individuals (computed in the same way as the starting population). Parent individuals for crossover are randomly selected from the elite and non-elite set, i.e. one from each pool. The crossover operator randomly computes weights for each patient category individually, by using a Gaussian distribution with mean at the center of the two parent weights for that particular patient category and a standard deviation chosen as the square of the distance of the two weights divided by 3, see Figure \ref{fig:GeneticAlgorithm}. Further, each offspring is subjected to a mutation operator with respect to a chosen mutation probability. 

The fitness of newly generated individuals is evaluated using the simulation model, as described above. Finally, all individuals of the new population are sorted with respect to their fitness value and the elite and non-elite sets are updated.

To avoid getting stuck in a local minimum, i.e. low heterogeneity in the population, a diversification step is included in the algorithm. The diversification step is applied if two conditions are satisfied: (1) the fitness values of the dominant half of the population (i.e. with lower fitness values) varies less than 1\% and (2) during the last 50 generations no improvement of the overall best solution was achieved.

As mentioned above, the generated training instances build the basis for the computation of objective values within the genetic algorithm. In particular, to compute the fitness of a weight vector $\beta$, all 10,000 training instances are simulated and the resulting patient data sets are aggregated to a single data set that is then used to compute the corresponding objective value.

The genetic algorithm is parametrized as follows. The population size is chosen to be 80 individuals, out of which 20 form the elite set of each generation. Of the remaining 60 individuals, 50 are generated by the crossover operator while 10 are randomly generated. The mutation probability $p_{mut}$ for each offspring is set to 0.75, i.e. with a 75\% chance weight is shifted from $\beta_i$ to $\beta_j$ where $i$ and $j$ are randomly selected. The share of weight $\beta_i$ that is shifted, is uniformly chosen between 0 and 50\%. 

After each generation of a new weight vector $\beta$ (either by mutation or crossover), $\beta$ is corrected to satisfy two pre-defined conditions. First, the sum of all weights must be 10 (to ensure comparability). Second, weights per consultation type must be strictly decreasing with respect to patient priorities, i.e. $b_{i,k} > b_{j,k}$ for $i<j$ and $k=1,2$. Note that randomly generated weight vectors are also required to satisfy those constraints.

Finally, the number of generations is chosen as 200. The resulting weights per scenario and objective functions are summarized in Table \ref{tab::APQWeights}.

\begin{table}[]
\begin{center}
\small
\begin{tabular}{cc cccc |c| cccc |c}
\hline
\multicolumn{1}{l}{Scenario} & Objective & $\beta_{2,1}$ & $\beta_{3,1}$ & $\beta_{4,1}$ & $\beta_{5,1}$ & $\sum \beta_{j,1}$ & $\beta_{2,2}$ & $\beta_{3,2}$ & $\beta_{4,2}$ & $\beta_{5,2}$ &  $\sum \beta_{j,2}$\\
\hline
\multirow{4}{*}{Base}        & TTDL      & 6.727         & 2.587         & 0.480         & 0.174           & 9.967           & 0.021         & 0.008         & 0.003         & 0.001           & 0.033           \\
                             & C-30      & 6.606         & 2.522         & 0.558         & 0.199           & 9.885           & 0.035         & 0.032         & 0.032         & 0.016           & 0.115           \\
                             & C-15      & 6.737         & 2.438         & 0.524         & 0.187           & 9.886           & 0.041         & 0.033         & 0.030         & 0.011           & 0.114           \\
                             & TWT       & 3.335         & 0.157         & 0.060         & 0.016           & 3.568           & 3.421         & 1.853         & 0.579         & 0.579           & 6.432           \\
                             \hline
HR                    & \multirow{6}{*}{C-15}    & 7.400  & 2.086  & 0.392  & 0.006  & 9.884 & 0.035  & 0.029  & 0.026  & 0.026  & 0.116 \\
LV                    &                          & 6.375  & 2.720  & 0.548  & 0.211  & 9.855 & 0.055  & 0.037  & 0.037  & 0.016  & 0.145 \\
HV                    &                          & 7.206  & 2.184  & 0.409  & 0.124  & 9.923 & 0.026  & 0.022  & 0.018  & 0.011  & 0.077 \\
U                     &                          & 6.8124 & 2.2876 & 0.5341 & 0.2124 & 9.847 & 0.0713 & 0.0332 & 0.0246 & 0.0245 & 0.154 \\
T3                    &                          & 8.016  & 1.732  & 0.213  & 0.023  & 9.984 & 0.007  & 0.003  & 0.003  & 0.003  & 0.016 \\
T5                    &                          & 5.4179 & 2.8171 & 0.4584 & 0.2023 & 8.896 & 0.814  & 0.1625 & 0.082  & 0.0457 & 1.104                  \\
                             \hline
\end{tabular}   
\caption{\ac{APQ} Weights for Different Scenario and Objective Function Combinations Computed by the Genetic Algorithm}
\label{tab::APQWeights}
\end{center}
\end{table}

For the base scenario, one may observe that weights for the first consultation are relatively similar if the objective function includes the \ac{TTD} target. In particular, weights for category 2 patients are approximately 2.5 to 3 times higher than those for category 3 patients, which are 4.5 to 5.5 times higher than those for category 4 patients. Interestingly, this changes for the TWT objective, where category 3 patients have significantly less weight compared to other first consultation weights. As a result, category 3 patients may less frequently ``overtake'' category 2 patients and more frequently be ``overtaken'' by category 4 patients. 

As could have been expected, the sum of weights assigned to second consultation queues is significantly lower for the TTDL objective and significantly higher for the TWT objective compared to combined measures.

The weights for scenarios with either altering variance or higher arrival rate show a similar behavior to the weights for the base scenario an the C-15 objective. Only for the HR (increased arrival rate) scenario is the $\beta_{5,1}$ weight significantly lower. This may result from the observation that for a system under severe stress, i.e. longer queue lengths, it is harder to meet \ac{TTD} targets for patients of higher priority.

The weight structure significantly changes for the T3 scenario (a peak of category 3 patients). Interestingly, the weight for the first consultation of category 2 patients is approximately 4.6 times higher than for category 3 patients. The large number of category 3 patients will most likely result in longer waiting times of those patients. Hence, the larger gap between weights may prevent too frequent overtaking of category 3 patients. On the other side, $\beta_{3,1}$ is approximately 8 times higher than $\beta_{4,1}$ for scenario T3. This may be also explained by more crowded $Q_{3,1}$ queues in this scenario, compared to the base scenario. The sum of weights for the second consultation is significantly lower compared to other configurations, as a larger number of ``higher'' priority patients negatively impacts the \ac{TTD} rates more than the overall waiting time.       
As would be expected, weight distributions for the T5 scenario assign higher weights to category 5 patients. In addition, the sum of weights for the second consultation is higher than for all other scenarios based on the C-15 objective. An intuitive explanation might be that it is necessary to reduce the waiting times for the second consultation of higher priority patients, in order to limit the waiting times for category 5 patients.

\subsection{Feature Engineering}

To train an \ac{ML} model for patient selection from decisions made by near optimal schedules of \ac{ED} instances, one is required to define a state representation of the \ac{ED} at the time a decision was made. This state representation can then be used as a set of features to form a sample for the \ac{ML} model. This sample is then composed by the feature set and the recorded response, i.e. the (near) optimal decision at the considered point in time $t$. To formally describe the set of features, let $P_q$ be the set of patients waiting in $q \in \{ Q_{2,1} , Q_{3,1} , Q_{4,1} , Q_{5,1} , Q^d_{2,2} , Q^d_{3,2} , Q^d_{4,2} , Q^d_{5,2} \}$ for doctor $d$ and, let $p_q^1$, $p_q^l$ and $p_q^m$ be the first, last and median patient waiting in queue $q$. Further, for a patient $p$ waiting in queue $q$ for first consultation, let $\hat{ttd}^p$ be the ``current'' time to doctor at time $t$, i.e. $\hat{ttd}^p = t - t_a^p$, as the real $ttd^p$ is only known upon the selection of patient $p$ for first consultation by doctor $d$. Waiting time measures $\hat{W}_1^p$ and $\hat{W}^p$ are defined accordingly as the waiting time that has been accumulated up to time $t$. Finally, let $I_X(x)$ be the index function that is $1$ if $x \in X$ and $0$ otherwise.

The set of feature may be divided into the following classes: general features, features concerning queues for the first consultation, feature concerning queues for the second consultation, and features concerning the register, triage and diagnostic activities. The entire feature set is outlined in Table \ref{tab::Features}.

\begin{table}[ht!]
	\begin{center}
		\begin{tabular}{l p{10cm} }
			\hline
			 Definition & Description \\
			 \hline
			 \multicolumn{2}{c}{General Features} \\
			 \hline
			 $t$ & Time of the day the decision was made \\
			 $d$ & Identifier of doctor \\
			 $|D_{idle}|$ & Number of idle doctors at time $t$ \\
			 $ | \{ p \in P_j | t_a^p \leq t \wedge \hat{ttd}^{p} \geq t_{ttd}^j \}| $  & Number of category $j$ patients that arrived before $t$ and missed the target time ($j \in J$) \\
			 $ | \{ p \in P_j | t_a^p \leq t \}| $  & Number of category $j$ patients that arrived before $t$ ($j \in J$) \\
			\hline
			\multicolumn{2}{c}{First Consultation Queue Features for queue $Q_{j,1}, j \in J$} \\
			 \hline
			 $|P_{Q_{j,1}}|$ & The length of queue $Q_{j,1}$ \\
			 $\hat{ttd}^{p_{Q_{j,1}}^1}$ & The current time to doctor of the first patient in queue $Q_{j,1}$ \\
			 $\hat{ttd}^{p_{Q_{j,1}}^l}$ & The current time to doctor of the last patient in queue $Q_{j,1}$ \\
			 $\hat{ttd}^{p_{Q_{j,1}}^m}$ & The current time to doctor of the median patient in queue $Q_{j,1}$ \\
			 $\sum_{p \in P_{Q_{j,1}}} \hat{ttd}^{p} / |P_{Q_{j,1}}|$ & The average current time to doctor of patients in queue $Q_{j,1}$ \\
			 $|P_{Q_{j,1}}|$ & The length of queue $Q_{j,1}$ \\
			 $\hat{W}_1^{p_{Q_{j,1}}^1}$ & The total waiting time of the first patient in queue $Q_{j,1}$ \\
			 $\hat{W}_1^{p_{Q_{j,1}}^l}$ & The total waiting time of the last patient in queue $Q_{j,1}$ \\
			 $\hat{W}_1^{p_{Q_{j,1}}^m}$ & The total waiting time of the median patient in queue $Q_{j,1}$ \\
			 $\sum_{p \in P_{Q_{j,1}}} \hat{W}_1^{p} / |P_{Q_{j,1}}|$ & The average total waiting time of patients in queue $Q_{j,1}$ \\
			 $\sum_{p \in P_{Q_{j,1}}} I_{[t_{ttd}^j, \infty)}(\hat{ttd}^{p})$ & Number of patients in queue that missed the target time to doctor \\
			 $\sum_{p \in P_{Q_{j,1}}} I_{[t_{ttd}^j - 5, t_{ttd}^j)}(\hat{ttd}^{p})$ & Number of patients that need to be seen in the next 5 minutes to meet the target \\
			 $\sum_{p \in P_{Q_{j,1}}} I_{[t_{ttd}^j - 10, t_{ttd}^j - 5)}(\hat{ttd}^{p})$ & Number of patients that need to be seen between 5 and 10 minutes to meet the target \\
			 $\sum_{p \in P_{Q_{j,1}}} I_{[t_{ttd}^j - 15, t_{ttd}^j - 10)}(\hat{ttd}^{p})$ & Number of patients that need to be seen between 10 and 15 minutes to meet the target \\
			 $\sum_{p \in P_{Q_{j,1}}} I_{[t_{ttd}^j - 20, t_{ttd}^j - 15)}(\hat{ttd}^{p})$ & Number of patients that need to be seen between 15 and 20 minutes to meet the target, for $j\geq 3$ \\
			 $\sum_{p \in P_{Q_{j,1}}} I_{[t_{ttd}^j - 30, t_{ttd}^j - 20)}(\hat{ttd}^{p})$ & Number of patients that need to be seen between 20 and 30 minutes to meet the target, for $j\geq 4$ \\
			 $\sum_{p \in P_{Q_{j,1}}} I_{[t_{ttd}^j - 40, t_{ttd}^j - 30)}(\hat{ttd}^{p})$ & Number of patients that need to be seen between 30 and 40 minutes to meet the target, for $j\geq 4$ \\
			 \hline
			    \multicolumn{2}{c}{Second Consultation Queue Features for queue $Q^d_{j,2}, j \in J$} \\
			 \hline
			  $\hat{W}^{p_{Q^d_{j,2}}^1}$ & The total waiting time of the first patient in queue $Q^d_{j,2}$ \\
			 $\hat{W}^{p_{Q^d_{j,2}}^l}$ & The total waiting time of the last patient in queue $Q^d_{j,2}$ \\
			 $\hat{W}^{p_{Q^d_{j,2}}^m}$ & The total waiting time of the median patient in queue $Q^d_{j,2}$ \\
			 $\sum_{p \in P_{Q^d_{j,2}}} \hat{W}^{p} / |P_{Q^d_{j,2}}|$ & The average total waiting time of patients in queue $Q^d_{j,2}$ \\
			 \hline
			    \multicolumn{2}{l}{Register, Triage and Diagnostic Features} \\
			 \hline
			 $|P_{Test}|$ & Number of patients at diagnostics at time $t$ \\
			 $|P^d_{Test}|$ & Number of patients at diagnostics at time $t$ that were first seen by doctor $d$ \\
			 $|P_{Register, Triage}|$ & Number of patients either waiting for or are in Register/Triage activities  at time $t$ \\
			 \hline
		\end{tabular}
		\caption{Summary of Feature Definitions}
		\label{tab::Features}
	\end{center}
\end{table}

\end{document}